\documentclass[a4paper]{article}
\usepackage{amsmath,amssymb}
\usepackage{graphicx}
\usepackage{subcaption}
\usepackage[margin=1in]{geometry}
\usepackage{hyperref}
\usepackage{float}
\usepackage{mdframed}
\usepackage{xcolor}
\usepackage{eso-pic}
\usepackage{pgfplots}
\usepackage{comment}
\pgfplotsset{compat=1.18}
\usepackage{colortbl}
\usepackage{booktabs}
\usepackage{tikz}
\usepackage{makecell}
\usepackage{natbib}
\usepackage{threeparttable}
\usepackage{multirow}
\usepackage{placeins}
\usepackage{diagbox}
\usepackage[most]{tcolorbox}
\newcommand{\radar}{0.19}
\newtcolorbox{promptbox}{
  enhanced,
  breakable,
  colback=gray!8,
  colframe=black,
  boxrule=0.4pt,
  sharp corners,
  left=6pt,
  right=6pt,
  top=6pt,
  bottom=6pt,
  overlay first={
    \draw[gray!8,line width=1.2pt]
      ([yshift=-0.2pt]frame.south west) --
      ([yshift=-0.2pt]frame.south east);
  },
  overlay middle={
    \draw[gray!8,line width=1.2pt]
      ([yshift=0.2pt]frame.north west) --
      ([yshift=0.2pt]frame.north east);
    \draw[gray!8,line width=1.2pt]
      ([yshift=-0.2pt]frame.south west) --
      ([yshift=-0.2pt]frame.south east);
  },
  overlay last={
    \draw[gray!8,line width=1.2pt]
      ([yshift=0.2pt]frame.north west) --
      ([yshift=0.2pt]frame.north east);
  }
}

\begin{document}

\AddToShipoutPictureFG*{%
  \AtPageLowerLeft{%
    \raisebox{3cm}{%
      \hspace{0.5cm}%
    }%
  }%
}

\title{Personality, Role, and Expressive Style in Large Language Models: An Interactionist Analysis}
\author{
Moe Nagao$^{1,2}$ \quad Koichiro Terao$^{1,2}$ \quad Mikio Nakano$^{3}$ \quad Naoto Iwahashi$^{1,2}$\\[2mm]
$^{1}$Okayama Prefectural University ~ $^{2}$AI \& Humans Lab ~ $^{3}$C4A Research Institute\\
}
\date{}
\maketitle
\begin{abstract}
Prompt-based personality control is a key technique for designing large language model (LLM) dialogue agents that behave consistently and appropriately across social contexts. However, specifying Big Five personality traits (BFTs) in a prompt does not necessarily ensure that the intended traits are expressed in generated utterances. This paper investigates this mismatch from an interactionist perspective, viewing personality expression as a context-dependent outcome shaped by the interplay between trait specification and situational factors. Specifically, we analyze how perceived BFT expression in LLM-generated dialogue is influenced by three prompt factors: assigned personality traits, dialogue roles, and expressive styles. Using a factorial design that systematically combines six personality conditions, three role conditions, and three expressive-style conditions, we generate 1,080 LLM-agent dialogues in each of English and Japanese. We then evaluate the target agent's utterances using an LLM-as-a-judge framework to estimate expressed Big Five traits. The results show that expressed personality is shaped not only by explicit trait specification, but also by dialogue role and expressive style. These effects are trait-specific: dialogue role strongly influences Openness, expressive style substantially shapes Conscientiousness and Agreeableness, and explicit trait specification dominates Neuroticism. Moreover, even in the absence of explicit personality-trait specification, social and expressive conditions induce distinct personality-like impressions. 
Cross-linguistic comparisons show broadly similar personality-expression patterns between English and Japanese dialogues, with noticeable differences emerging only under specific combinations of personality, role, and expressive style.
These findings suggest that personality control in LLM agents should be understood not as a direct consequence of trait prompting, but as a context-dependent process involving the interaction of personality specification, social role, and expressive style.
\end{abstract}

\section{Introduction}

\begin{figure}[b]
    \centering
    \includegraphics[width=0.84\linewidth]{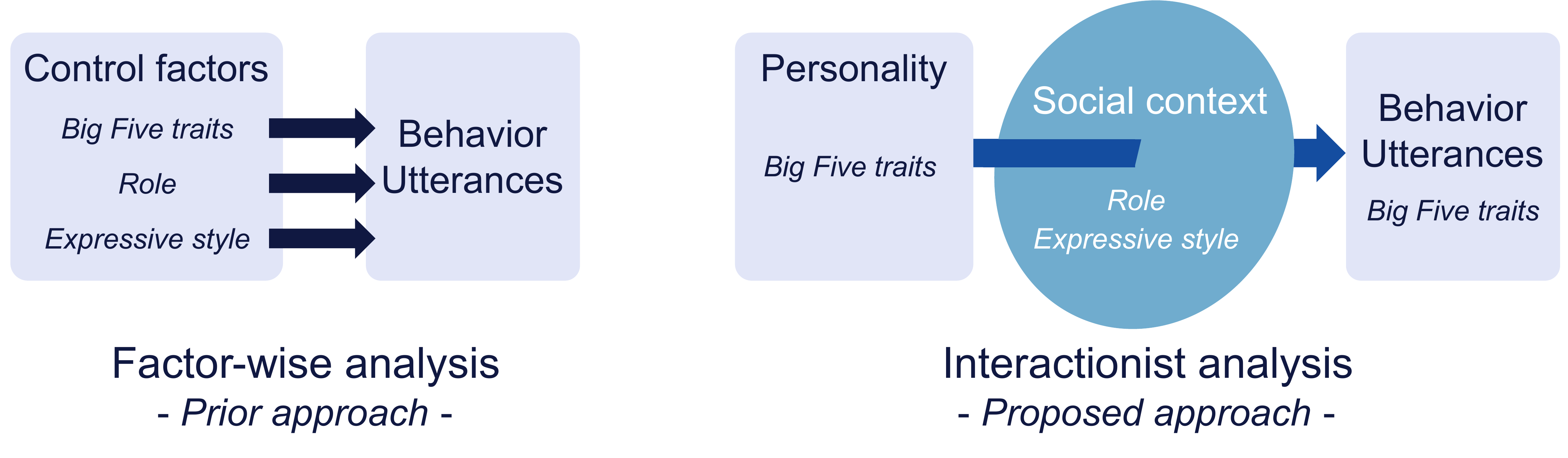}
    \vspace{-2mm}
    \caption{Overview of our interactionist analysis framework. Unlike prior factor-wise analyses that examine control factors in isolation, our approach analyzes how internally specified Big Five traits interact with dialogue-level social context, including role and expressive style, to shape generated utterances.}
    \label{fig:interactionist_analysis}
\end{figure}

Large language models (LLMs) are increasingly used as dialogue agents whose behavior must be controlled across diverse social contexts. 
One important direction is to endow such agents with human-like personality traits so that they can produce responses that are consistent, appropriate, and interpretable. 
Prior work has therefore explored prompt-based methods for controlling LLM agents by specifying personality-related attributes, including personas, Big Five traits (BFTs)~\citep{goldberg1990bigfive}, roles~\citep{kong2024better, cao2026biased}, and emotions~\citep{ramirez2023controlling, jiang2023evaluating}. 
However, recent studies have shown that prompt-based personality modulation does not always lead to the intended behavioral changes, and that internally specified BFTs may not be expressed in generated utterances as intended~\citep{zakazov2024assessing,nagao2025sag}. 
These findings reveal a central challenge for controllable LLM agents: specifying a personality trait in a prompt does not necessarily ensure that the corresponding trait will be expressed in dialogue.

A possible reason for this mismatch is that personality expression is shaped not only by explicit trait specification, but also by the social conditions under which dialogue takes place. 
For example, an agent instructed to be open-minded may express this trait differently when engaging in casual chat than when acting as a salesperson in a task-oriented interaction. 
Similarly, an emotional or rational expressive style may change how the same personality prompt is perceived. 
Although prior studies have examined persona prompting, intrinsic personality traits, roles, and emotional expression in LLMs, these factors have largely been analyzed in isolation. 
As a result, it remains unclear how assigned personality traits are expressed under different combinations of dialogue roles and expressive styles.

In this paper, we address this issue by analyzing personality expression in LLM-generated dialogue from an interactionist perspective. 
As illustrated in Figure~\ref{fig:interactionist_analysis}, our approach contrasts with prior factor-wise analyses that examine control factors separately.
Specifically, we analyze how internally specified personality traits interact with two dialogue-level social factors: the agent's role and expressive style.
Our analysis is grounded in interactionist theories of personality~\citep{sheldon1997trait, wood2006cross, MISCHEL2009282, fleeson2001toward, mischel2013personality, fleeson2015whole}, which view behavior not simply as a direct manifestation of internal traits, but as an outcome of interactions between trait dispositions and social context. 
From this perspective, personality expression in LLM dialogue may depend not only on which traits are specified, but also on the role assigned to the agent and the expressive manner in which it is instructed to respond.

We operationalize personality expression using the Big Five framework, one of the most widely used models of personality, comprising Openness, Conscientiousness, Extraversion, Agreeableness, and Neuroticism. 
Using a factorial design, we systematically combine six personality conditions, three role conditions, and three expressive-style conditions, and generate dialogues between LLM agents in both English and Japanese. 
The target agent's utterances are then evaluated using an LLM-as-a-judge framework to estimate expressed Big Five traits. 
This design allows us to examine not only whether explicit trait prompts affect expressed personality, but also how their effects are modulated by dialogue roles and expressive styles, and whether these patterns are broadly shared across languages.

Our results show that personality expression in LLM-generated utterances should be understood not only as a consequence of explicit trait prompts, but also as an outcome shaped by dialogue-level social factors. 
This paper makes the following contributions:

\begin{figure*}[t]
    \centering
    \includegraphics[width=0.96\linewidth]{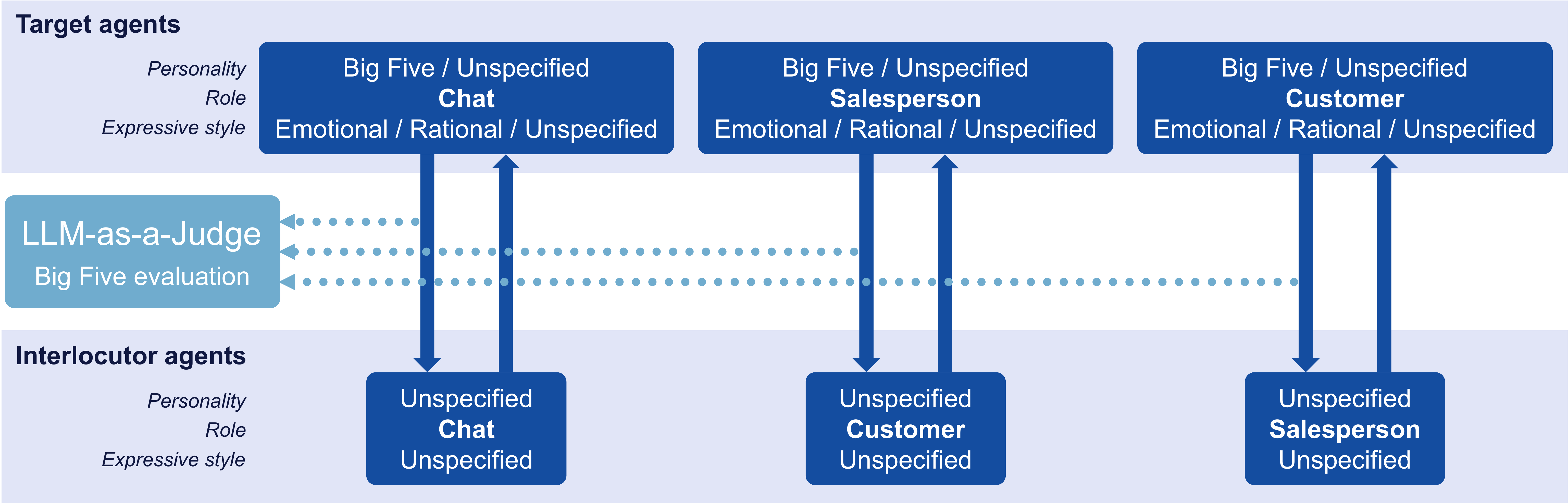}
    \caption{Overview of the analysis method.}
    \label{analysis_figure}
\end{figure*}

\begin{enumerate}
    \item
        We introduce an interactionist framework for analyzing personality expression in LLM-generated dialogue, focusing on the joint effects of prompt-specified Big Five traits, dialogue roles, and expressive styles.
    \item
        We provide a systematic factorial analysis showing that personality expression is trait-specific and context-dependent: the relative influence of personality specification, role, and expressive style differs substantially across Big Five dimensions.
    \item
        We show that role and expressive style shape expressed Big Five traits even in the absence of explicit personality-trait specification, indicating that personality-like impressions can emerge from dialogue-level social and expressive conditions.
    \item
        We compare English and Japanese dialogues and find broadly similar personality-expression patterns, with noticeable differences emerging under specific combinations of personality, role, and expressive style.
\end{enumerate}

Taken together, our findings suggest that personality expression in LLMs can be usefully analyzed from an interactionist perspective: expressed traits are not simply direct reflections of internal trait specifications, but are shaped by the social conditions in which dialogue takes place. 
This underscores the importance of situating personality control in LLMs within broader social and linguistic contexts.

The remainder of this paper is organized as follows. Section~\ref{sec:method} describes our analysis methodology. Section~\ref{sec:results} presents the results. Section~\ref{sec:discussion} discusses the findings. Section~\ref{sec:related_work} reviews related work. Finally, Section~\ref{sec:conclusion} concludes the paper.

\section{Method}
\label{sec:method}

Figure~\ref{analysis_figure} provides an overview of our analysis method.
We first generate dyadic dialogues between pairs of a target LLM agent and an interlocutor LLM agent.
We then analyze how the prompt configurations influence the generated dialogue.
For each dialogue, an LLM judge evaluates the extent to which the target agent expresses each Big Five trait using a five-point Likert scale.

\subsection{Prompt Factors}

We manipulate three prompt factors assigned to the target agent: personality condition, role, and expressive style.
Taken together, these factors allow us to examine how explicitly specified personality traits, dialogue roles, and expressive styles shape conversational behavior.
The interlocutor agent is assigned only a role, allowing us to examine how the target agent's assigned attributes manifest across different conversational contexts.

\paragraph{Personality.}
To examine how explicitly specifying a Big Five trait shapes the generated dialogue, we define six personality conditions.
The Unspecified condition serves as the baseline, in which no Big Five trait is specified.
In each of the five trait-specific conditions, one Big Five trait is specified as high, while the remaining traits are left unspecified.

Specifically, we consider the following six personality conditions:
\begin{enumerate}
  \item Unspecified: no Big Five trait is specified.
  \item Extraversion: high Extraversion is specified.
  \item Agreeableness: high Agreeableness is specified.
  \item Conscientiousness: high Conscientiousness is specified.
  \item Neuroticism: high Neuroticism is specified.
  \item Openness: high Openness is specified.
\end{enumerate}

\paragraph{Role.}
We consider three roles to examine how personality-related expressions vary between unconstrained and socially constrained interactional settings.
The Chat role does not impose a specific task objective and is therefore treated as the role baseline.
We select the Salesperson role as a representative example of a socially constrained role and include the Customer role as the counterpart in the Salesperson–Customer pairing.

Specifically, we consider the following three roles:
\begin{enumerate}
  \item Chat: a chat partner.
  \item Salesperson: a salesperson at a consumer electronics retail store.
  \item Customer: a customer visiting a consumer electronics retail store.
\end{enumerate}
The resulting role pairings are Chat--Chat, Salesperson--Customer, and Customer--Salesperson.

\paragraph{Expressive Style.}
We define three Expressive Style conditions to examine how explicitly specifying an expressive orientation influences the generated dialogue. The Unspecified condition serves as the expressive-style baseline, in which no expressive style is specified. In the remaining conditions, either an emotional or a rational expressive style is specified. 
We operationalize expressive style using two prompt-based conditions, Emotional and Rational, which contrast affectively oriented and cognitively oriented forms of expression. This contrast is motivated by established theoretical frameworks, including the stereotype content model in social cognition~\citep{fiske2002model} and dual-process theory in cognitive psychology~\citep{kahneman2011thinking}, both of which distinguish affective from cognitively oriented aspects of social expression.
Specifically, we consider the following three expressive style conditions:

\begin{enumerate}
  \item Unspecified: no expressive style is specified.
  \item Emotional: an emotionally expressive style is specified.
  \item Rational: a rationally expressive style is specified.
\end{enumerate}

\paragraph{Prompt design.}
The prompt formulations used to operationalize the Big Five traits follow \citet{jiang2023evaluating} and are provided in Appendix~\ref{appendix_prompt_bigfive}. These formulations use the same trait-descriptive adjectives as \citet{jiang2023evaluating}, originally taken from \citet{mccrae1992introduction}. The prompt formulations for expressive styles are provided in Appendix~\ref{appendix_prompt_express}. An example prompt for the Salesperson role in an English dialogue under the Extraversion and Emotional conditions is provided in Appendix~\ref{appendix_prompt_salesperson}.

\subsection{LLM-based Personality Evaluation}

We adopt an LLM-as-a-judge paradigm for personality assessment. 
Our goal is not to measure the target agent's ``true'' personality, but to estimate the Big Five traits perceived from its generated utterances. 
This use of LLM-based evaluation is motivated by prior work showing that LLM-based personality judgments can align with human personality judgments at levels comparable to, or exceeding, ratings by familiar human informants~\citep{llm_as_a_judge}. 
Although our setting differs in that we assess personality traits expressed in dialogue rather than self-reported personality, this evidence supports the use of LLM-based ratings as a scalable proxy for human judgments of perceived personality expression.

We therefore interpret the resulting scores as estimates of perceived personality expression, rather than as direct psychometric measurements. 
The prompt used for personality evaluation is presented in Appendix~\ref{appendix_prompt_evaluation}. 
To assess the robustness of these estimates to evaluator choice, we compare ratings from two independent evaluator models in Section~\ref{sec:evaluator_evaluation}.

% We adopt an LLM-as-a-judge paradigm for personality assessment. 
% This choice is motivated by prior work showing that LLM-based personality judgments can align with human self-reports at levels comparable to, or exceeding, ratings by family members and friends~\citep{llm_as_a_judge}. 
% Although our setting differs in that we assess personality traits expressed in dialogue rather than self-reported personality, this evidence supports the use of LLMs as scalable evaluators of personality-related cues in text. 
% We therefore treat LLM-based ratings as scalable estimates of perceived personality expression, rather than as replacements for full human validation. The prompt used for personality evaluation is presented in
% Appendix~\ref{appendix_prompt_evaluation}.
% To further assess the robustness of these estimates, we compare ratings from two independent evaluator models in Section~\ref{sec:evaluator_evaluation}.

\subsection{Analysis Setup}

For dialogue generation, we used GPT-5.2 (gpt-5.2-2025-12-11), and for dialogue evaluation, we used Gemini 2.5 Flash (gemini-2.5-flash).
We used a separate evaluation model to mitigate self-preference bias~\citep{NEURIPS2023_91f18a12}, whereby a model may assign overly favorable judgments to outputs resembling its own generation style.

Each dialogue consisted of 10 turns.
For each combination of role, personality condition, and expressive style condition, we generated 20 dialogues.
Dialogues were generated in both English and Japanese, whereas the evaluation prompt was written in English and applied uniformly to dialogues in both languages.

The evaluation focused exclusively on the target agent. For each setting, Big Five trait scores were assessed for the target agent in each of the 20 generated dialogues. The means of the resulting scores were visualized using radar charts. Neuroticism scores were converted into Emotional Stability scores by subtracting them from 6 prior to visualization. This transformation ensures consistent axis directionality, such that higher values indicate more positive trait expressions across all axes.
For the ANOVA analyses, each generated dialogue was treated as one observation, rather than aggregating scores at the condition level. Thus, for each two-way ANOVA, we analyzed $2 \times 3 \times 20 = 120$ observations, yielding 114 residual degrees of freedom. For each three-way ANOVA, we analyzed $2 \times 3 \times 3 \times 20 = 360$ observations, yielding 342 residual degrees of freedom. These dialogue-level observations correspond to repeated stochastic generations under the same prompt condition, and the resulting statistical tests should therefore be interpreted as analyses of variation across generated dialogue samples rather than as measurements from independent human participants.

We analyze the effects of personality condition, role, and expressive style on Big Five trait expression in LLM agents.
Specifically, we conducted the following analyses:

\begin{description}
    \item [Effect of Personality.]
        To examine the direct effect of explicit Big Five trait specification, we evaluated Big Five trait expression under the Chat role, which serves as the role baseline, and the Unspecified expressive-style condition.
    \item [Joint Effects of Personality and Role.]
        To examine the joint effects of Personality and Role, we compared Big Five trait scores under the Unspecified expressive-style condition across Personality and Role conditions.
    \item [Joint Effects of Personality and Expressive Style.]
        To examine the joint effects of Personality and Expressive Style, we compared Big Five trait expression across combinations of Personality and Expressive Style conditions, under the Chat role condition.
    \item [Joint Effects of Personality, Role, and Expressive Style.]
        To examine the joint effects of Personality, Role, and Expressive Style, we compared Big Five trait expression scores across all combinations of these conditions.
    \item [Comparison Between English and Japanese Dialogues.]
        To examine cross-linguistic differences, we compared the evaluation results for English and Japanese dialogues under corresponding experimental conditions.
\end{description}

\section{Results}
\label{sec:results}

\subsection{Robustness of LLM-based Evaluation}
\label{sec:evaluator_evaluation}

Before presenting the main results, we examined whether the estimated Big Five scores were robust to the choice of LLM-based evaluator. 
In addition to the primary evaluator, Gemini 2.5 Flash, we re-evaluated the English dialogues using OpenAI o3-mini. 
For each of the 54 Personality--Role--Expressive Style conditions, we computed the mean score for each Big Five trait over 20 dialogue samples for each evaluator, and calculated the RMSE between the two sets of condition-level means. 
We also computed Pearson's correlation coefficients at the individual-dialogue level.
As shown in Table~\ref{tab:rmse_between_llms}, RMSE values were relatively small across all five traits, and Pearson's correlations indicated moderate-to-strong agreement between the two evaluator models. 
These results suggest that the estimated trait scores were not strongly dependent on the particular LLM-based evaluator used in this study.

However, this analysis should be interpreted only as evidence of inter-evaluator consistency among LLM-based judges. It does not establish that the estimated scores are equivalent to human judgments of personality expression. Accordingly, we treat the resulting scores as LLM-based estimates of perceived personality expression, rather than as validated measurements of how human annotators would perceive the same dialogues.

The following subsections report the results obtained using Gemini 2.5 Flash.

\begin{table}[t]
\centering
\caption{RMSE and Pearson's correlation coefficients between Gemini 2.5 Flash and OpenAI o3-mini evaluations of Big Five trait scores. RMSE was computed using condition-level mean scores, whereas Pearson's $r$ was computed at the individual-dialogue level.}
\label{tab:rmse_between_llms}
\begin{tabular}{lcc}
\hline
Trait & RMSE & Pearson's $r$ \\
\hline
Openness & 0.437 & 0.759\\
Conscientiousness & 0.529 & 0.770 \\
Extraversion & 0.568 & 0.819 \\
Agreeableness & 0.295 & 0.504 \\
Neuroticism & 0.275 & 0.903 \\
\hline
\end{tabular}
\vspace{5mm}
\end{table}

%%%%%%%%%%%%%%%%%%%%%%%%%%%%%%%%%%%%%%%%%%%%%%%%%%%%%%%%%%%%%%%%%%%%%
\subsection{Effect of Personality Condition}

\subsubsection*{Mean Big Five Trait Scores}

Figure~\ref{BFT} shows the expressed Big Five trait scores in English dialogues for the Chat role under the Unspecified expressive-style condition. 
Figure~\ref{BFT-Unspecified} shows the scores under the Unspecified personality condition, which we regard as a baseline reflecting the default expressive characteristics of LLMs.

Compared with the Unspecified personality condition, specifying a Big Five trait generally increased the score of the corresponding expressed trait. 
This pattern was especially clear for Openness and Neuroticism, whereas it was less pronounced for Conscientiousness and Agreeableness. 
These results suggest that prompt-specified personality conditions can influence perceived Big Five trait expression, but that the strength of this influence varies across trait dimensions. 
These baseline scores are used in the following sections as reference points for interpreting the joint effects of Personality, Role, and Expressive Style.

\begin{figure*}[t]
\input{figure_set_allBFTs_Chat_StyleUnsp}
\end{figure*}

\begin{figure*}[t]
\input{figure_JointEffects_of_BFTandRole}
\end{figure*}

\subsection{Joint Effects of Personality and Role}

\subsubsection*{Mean Big Five Trait Scores} 

Figure~\ref{BFT-Role} presents the expressed Big Five trait scores in English dialogues under the Unspecified expressive-style condition, separately for each Personality and Role condition.
Each panel corresponds to one Personality condition, and the three lines indicate the Chat, Salesperson, and Customer roles.

The results show that role assignment shaped expressed Big Five traits even when no explicit Big Five trait was specified. 
Under the Unspecified Personality condition, Conscientiousness was higher in the Salesperson and Customer roles than in the Chat role, whereas Openness was higher in the Chat role than in the Salesperson and Customer roles. 
This pattern suggests that purchase-related interactions encouraged more goal-directed and responsibility-oriented responses, whereas unconstrained chat allowed more flexible and exploratory responses.

The effect of Role also depended on the Personality condition. 
Role-related differences were relatively small under the Unspecified, Extraversion, and Agreeableness conditions, but became more pronounced under the Openness, Conscientiousness, and Neuroticism conditions. 
These observations suggest that role assignment does not simply add a uniform contextual effect, but modulates trait expression differently depending on the specified Personality condition.

\subsubsection*{Two-Dimensional Configuration of Mean Big Five Trait Scores} 

\begin{figure*}[t]
    \centering
  \includegraphics[width=0.75\linewidth]{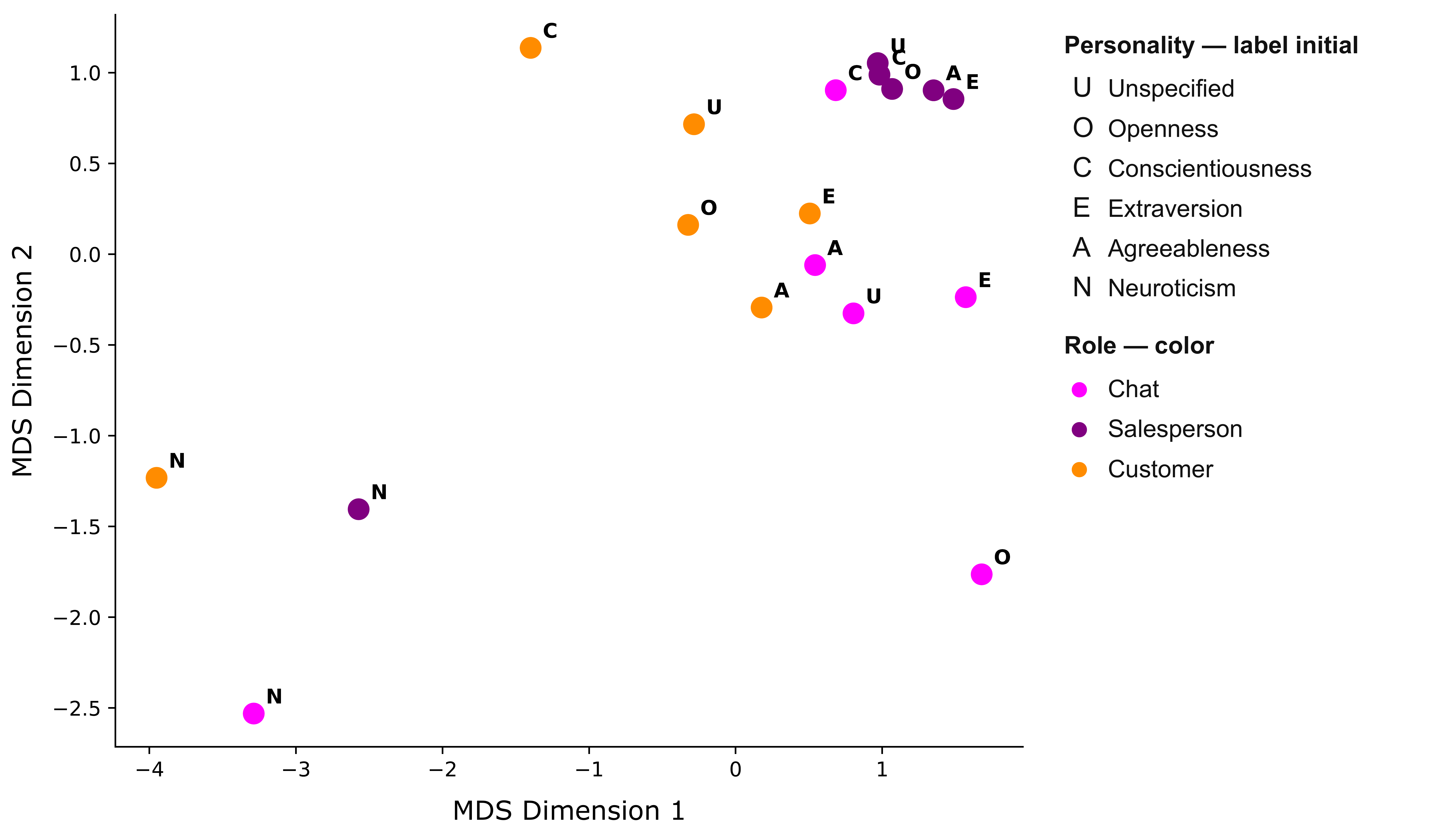}
    \caption{Two-dimensional configuration of the pentagonal mean Big Five score polygons in the radar charts for the joint effects of Personality and Role.}
  \label{fig:Personality_Role_MDS}
\end{figure*}

\begin{table}[th]
\centering
\caption{Two-way ANOVA results for the effects of Personality and Role on each expressed Big Five scale.}
\label{tab:Personality_Role_two_way_ANOVA}
\begin{tabular}{llrrrr}
\hline
Dependent Variable & Factor/Interaction & df & $F$ & $p$ & $\omega^2$ \\
\hline
\multirow{3}{*}{Openness}
 & Personality & 1, 114 & 153.82 & $< .001$ & 0.254 \\
 & Role & 2, 114 & 128.39 & $< .001$ & 0.424 \\
 & Personality $\times$ Role & 2, 114 & 37.60 & $< .001$ & 0.122 \\
\hline
\multirow{3}{*}{Conscientiousness}
 & Personality & 1, 114 & 33.63 & $< .001$ & 0.134 \\
 & Role & 2, 114 & 21.77 & $< .001$ & 0.171 \\
 & Personality $\times$ Role & 2, 114 & 25.31 & $< .001$ & 0.200 \\
\hline
\multirow{3}{*}{Extraversion}
 & Personality & 1, 114 & 75.49 & $< .001$ & 0.291 \\
 & Role & 2, 114 & 32.35 & $< .001$ & 0.245 \\
 & Personality $\times$ Role & 2, 114 & 0.51 & 0.600 & 0.000 \\
\hline
\multirow{3}{*}{Agreeableness}
 & Personality & 1, 114 & 8.41 & 0.0045 & 0.053 \\
 & Role & 2, 114 & 6.39 & 0.0023 & 0.077 \\
 & Personality $\times$ Role & 2, 114 & 2.35 & 0.0996 & 0.019 \\
\hline
\multirow{3}{*}{Neuroticism}
 & Personality & 1, 114 & 3258.50 & $< .001$ & 0.926 \\
 & Role & 2, 114 & 55.30 & $< .001$ & 0.031 \\
 & Personality $\times$ Role & 2, 114 & 16.63 & $< .001$ & 0.009 \\
\hline
\end{tabular}
\end{table}

Figure~\ref{fig:Personality_Role_MDS} shows a two-dimensional configuration of the pentagonal mean-score polygons in Figure~\ref{BFT-Role}, obtained by multidimensional scaling (MDS). 
The MDS solution yielded a normalized stress value of 0.055, indicating a reasonably good two-dimensional representation of the original dissimilarities.

The MDS plot provides an exploratory overview of how Personality and Role jointly shaped the overall configuration of expressed Big Five profiles. 
In particular, under the Salesperson role, the points corresponding to all Personality conditions except Neuroticism form a relatively tight cluster. 
This suggests that the Salesperson role constrained the overall personality-expression profile, reducing differences among Personality conditions except when Neuroticism was explicitly specified. 
Because the MDS analysis was used only for exploratory visualization, the primary statistical evidence is provided by the ANOVA results below.

\subsubsection*{Analysis of Variance}

Table~\ref{tab:Personality_Role_two_way_ANOVA} presents the results of two-way ANOVAs examining the effects of Personality and Role on each expressed Big Five trait. 
We conducted five separate analyses, one for each Big Five trait dimension as the dependent variable. 
In each analysis, Personality had two levels, Unspecified and the corresponding dimension-specific condition, and Role had three levels, Chat, Salesperson, and Customer.

The ANOVA results confirm that the relative contributions of Personality, Role, and their interaction differed substantially across Big Five dimensions. 
For Openness, Role showed the largest effect, $\omega^2 = 0.424$, exceeding the effect of Personality, $\omega^2 = 0.254$. 
This indicates that Openness expression was strongly shaped by whether the agent was placed in an unconstrained chat setting or in a task-oriented retail interaction. 
For Conscientiousness, the Personality $\times$ Role interaction was the largest effect, $\omega^2 = 0.200$, suggesting that the effect of explicit Conscientiousness specification depended on the assigned Role. 
For Extraversion, both Personality and Role had substantial main effects, but their interaction was negligible, indicating that the two factors contributed largely independently. 
For Agreeableness, the effects were statistically significant but relatively small. 
For Neuroticism, Personality had by far the largest effect, $\omega^2 = 0.926$, indicating that explicit Neuroticism specification dominated the expressed Neuroticism score, while Role-related effects were comparatively small.

Overall, these results show that Role is not merely a background condition but an active contextual factor that shapes perceived personality expression. 
However, the extent and form of this role-based modulation differ across Big Five dimensions.

\subsection{Joint Effects of Personality and Expressive Style}

\begin{figure*}[t]
\input{figure_set_JointEffects_of_BFTandExpressive}
\end{figure*}

\subsubsection*{Mean Big Five Trait Scores}

Figure~\ref{BFT-Style} presents the expressed Big Five trait scores in English dialogues under the Chat role condition, separately for each Personality and Expressive Style condition. 
Each panel corresponds to one Personality condition, and the three lines indicate the Unspecified, Emotional, and Rational expressive-style conditions.

The results show that Expressive Style systematically shaped perceived Big Five trait expression, even when no explicit Big Five trait was specified. 
Under the Unspecified Personality condition, the Emotional style increased Agreeableness and decreased Conscientiousness and Emotional Stability, whereas the Rational style increased Conscientiousness and decreased Extraversion and Openness. 
This pattern suggests that expressive-style instructions can induce personality-like impressions by making the agent's responses more affectively expressive under the Emotional condition and more cognitively oriented under the Rational condition.

The effects of Emotional and Rational styles also tended to operate in opposite directions relative to the Unspecified expressive-style baseline. 
Emotional expression generally shifted the profile toward higher Agreeableness and lower Emotional Stability, whereas Rational expression shifted the profile toward higher Conscientiousness and lower Extraversion. 

Together, these observations suggest that Expressive Style can function as a contextual factor that shapes perceived personality expression, rather than merely changing the surface form of responses.

\begin{figure*}[t]
  \centering
  \includegraphics[width=0.75\linewidth]{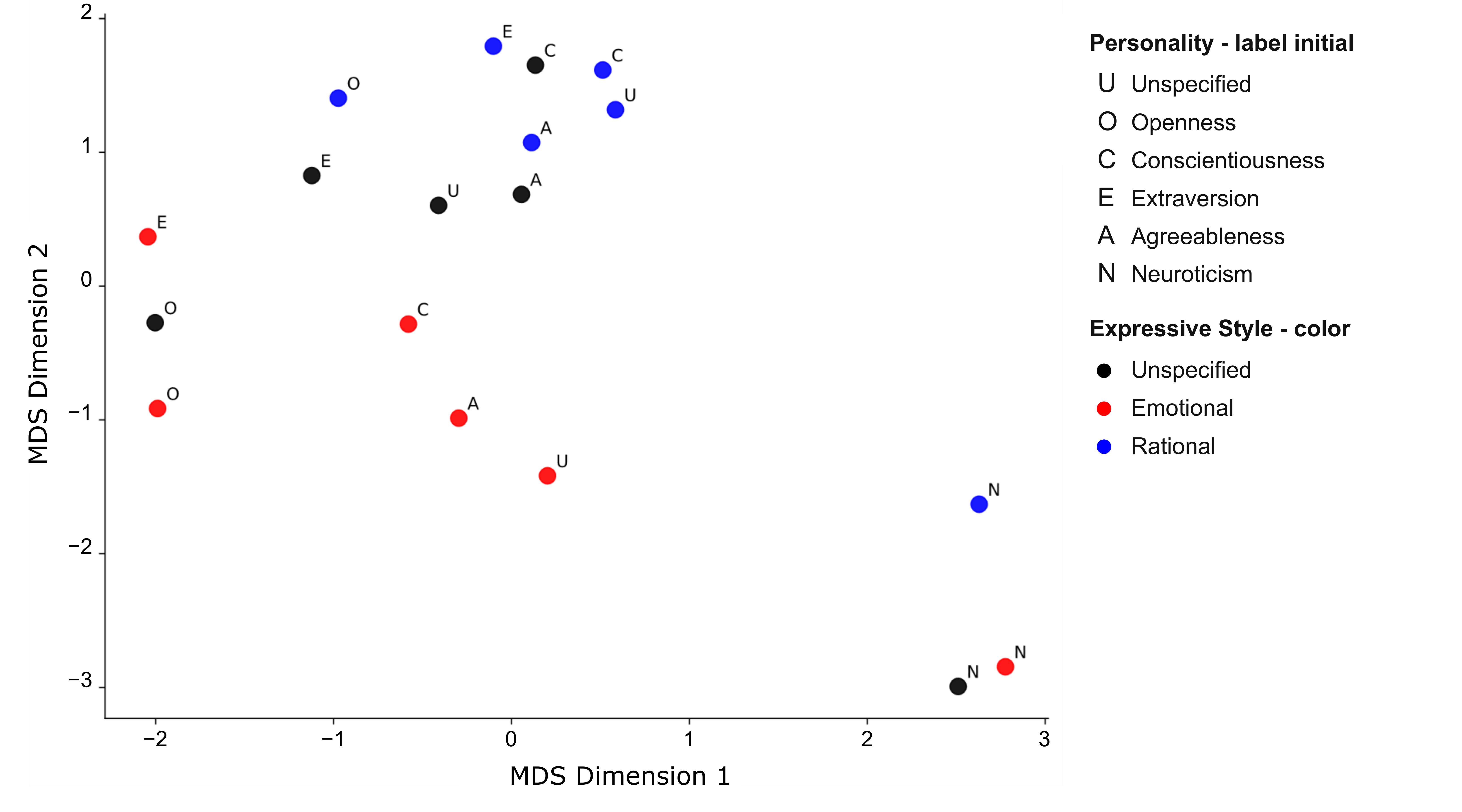}
  \caption{Two-dimensional configuration of the pentagonal mean Big Five score polygons in the radar charts for joint effects of Personality and Expressive Style.}
  \label{fig:Personality_Expressive_style_MDS}
\end{figure*}

\begin{table}[th]
\centering
\begin{threeparttable}
\caption{Two-way ANOVA results for the effects of Personality and Expressive Style on each expressed Big Five scale.}
\label{tab:anova_chat_personality_style}
\begin{tabular}{llrrrr}
\hline
Dependent Variable & Factor/Interaction & df & $F$ & $p$ & $\omega^2$ \\
\hline
\multirow{3}{*}{Openness}
& Personality & 1,114 & 416.72 & $< .001$ & 0.730 \\
& Expressive Style & 2,114 & 16.79 & $< .001$ & 0.055 \\
& Personality $\times$ Expressive Style & 2,114 & 2.17 & .119 & 0.004 \\
\hline
\multirow{3}{*}{Conscientiousness}
& Personality & 1,114 & 31.43 & $< .001$ & 0.065 \\
& Expressive Style & 2,114 & 149.27 & $< .001$ & 0.638 \\
& Personality $\times$ Expressive Style & 2,114 & 9.78 & $< .001$ & 0.038 \\
\hline
\multirow{3}{*}{Extraversion}
& Personality & 1,114 & 91.48 & $< .001$ & 0.329 \\
& Expressive Style & 2,114 & 28.09 & $< .001$ & 0.197 \\
& Personality $\times$ Expressive Style & 2,114 & 6.33 & .002 & 0.039 \\
\hline
\multirow{3}{*}{Agreeableness}
& Personality & 1,114 & 27.73 & $< .001$ & 0.095 \\
& Expressive Style & 2,114 & 64.70 & $< .001$ & 0.454 \\
& Personality $\times$ Expressive Style & 2,114 & 4.11 & .019 & 0.022 \\
\hline
\multirow{3}{*}{Neuroticism\tnote{a}}
& Personality & 1,114 & 699.58 & $< .001$ & 0.731 \\
& Expressive Style & 2,114 & 35.34 & $< .001$ & 0.072 \\
& Personality $\times$ Expressive Style & 2,114 & 35.34 & $< .001$ & 0.072 \\
\hline
\end{tabular}

\begin{tablenotes}
\footnotesize
\item[a] For Neuroticism, the main effect of Expressive Style and the Personality $\times$ Expressive Style interaction yielded identical values because the scores under the Neuroticism personality condition were fixed at the maximum value across all Expressive Style conditions. As a result, the Expressive Style effect was entirely attributable to variation within the Unspecified personality condition.
\end{tablenotes}
\end{threeparttable}
\end{table}

\subsubsection*{Two-Dimensional Configuration of Mean Big Five Trait Scores}

Figure~\ref{fig:Personality_Expressive_style_MDS} shows a two-dimensional configuration of the pentagonal mean-score polygons in Figure~\ref{BFT-Style}, obtained by multidimensional scaling (MDS). 
The MDS solution yielded a normalized stress value of 0.078, indicating a reasonably good two-dimensional representation of the original dissimilarities.

The MDS plot provides an exploratory overview of how Personality and Expressive Style jointly shaped the overall configuration of expressed Big Five profiles. 
Compared with the Unspecified expressive-style condition, the Emotional and Rational conditions tended to move the profiles in different directions in the two-dimensional space. 
This suggests that the two expressive styles induced distinct configurations of perceived personality expression. 
Because the MDS analysis was used only for exploratory visualization, the primary statistical evidence is provided by the ANOVA results below.

\subsubsection*{Analysis of Variance}

Table~\ref{tab:anova_chat_personality_style} presents the results of two-way ANOVAs examining the effects of Personality and Expressive Style on each expressed Big Five trait. 
We conducted five separate analyses, one for each Big Five trait dimension as the dependent variable. 
In each analysis, Personality had two levels, Unspecified and the corresponding dimension-specific condition, and Expressive Style had three levels, Unspecified, Emotional, and Rational.

The ANOVA results confirm that the relative contributions of Personality, Expressive Style, and their interaction differed substantially across Big Five dimensions. 
For Openness, Personality showed the largest effect, $\omega^2 = 0.730$, whereas the effect of Expressive Style was comparatively small, $\omega^2 = 0.055$, and the interaction was not significant. 
This indicates that Openness expression was primarily driven by explicit Openness specification rather than by expressive style.

For Conscientiousness and Agreeableness, Expressive Style showed the largest effects, $\omega^2 = 0.638$ and $\omega^2 = 0.454$, respectively. 
These results indicate that perceived Conscientiousness and Agreeableness were especially sensitive to whether the agent responded in a Rational or Emotional style. 
For Extraversion, both Personality and Expressive Style had substantial effects, suggesting that Extraversion expression was shaped jointly by explicit trait specification and expressive orientation.

For Neuroticism, Personality had the largest effect, $\omega^2 = 0.731$, but Expressive Style and the Personality $\times$ Expressive Style interaction also showed significant effects. 
As noted in Table~\ref{tab:anova_chat_personality_style}, the identical values for the Expressive Style main effect and the interaction reflect the fact that Neuroticism scores under the Neuroticism personality condition were fixed at the maximum value across expressive-style conditions. 
Thus, the observed expressive-style effect for Neuroticism was attributable to variation within the Unspecified personality condition.

Overall, these results show that Expressive Style can substantially alter perceived personality expression, but its influence is trait-specific. 
While explicit Personality specification dominated Openness and Neuroticism, Expressive Style played a central role in shaping Conscientiousness and Agreeableness.

%%%%%%%%%%%%%%%%%%%%%%%%%%%%%%%%%%%%%%%%%%%%%%%%%%%%%%%%%%%%%%%%%%%%%
\subsection{Joint Effects of Personality, Role, and Expressive Style}

\subsubsection*{Mean Big Five Trait Scores}

Table~\ref{tab:radar-personality-role-expressive-style} summarizes the expressed Big Five trait scores in English dialogues across Personality, Role, and Expressive Style conditions. Each row corresponds to a Personality condition, and each column corresponds to a Role condition. Within each panel, the three lines represent the Unspecified, Emotional, and Rational Expressive Style conditions.

The radar charts show that Expressive Style modulated the expression of prompt-specified traits in a role-dependent manner. Overall, Emotional and Rational styles did not shift the Big Five profiles uniformly; rather, their effects varied depending on both the specified Personality condition and the assigned Role. For example, across roles, the Emotional condition tended to enhance the expression of the specified trait under the Extraversion and Agreeableness conditions. In contrast, under the Conscientiousness condition, the Emotional style reduced the Conscientiousness score in the Chat role, whereas this reduction was not observed in the Salesperson and Customer roles. Similarly, under the Openness condition, the Emotional style increased the Openness score specifically in the Customer role, while the corresponding scores remained largely unchanged in the Chat and Salesperson roles.

These patterns indicate that Expressive Style can either facilitate or suppress the manifestation of a specified trait, depending on the social role in which the agent is placed.

\begin{table}[htbp]
\centering
\caption{Expressed Big Five trait scores across Personality, Role, and Expressive Style conditions.
Rows denote Personality conditions: the Unspecified condition and the five trait-specific conditions corresponding to the Big Five dimensions.
Columns denote dialogue roles: Chat, Salesperson, and Customer.
S $=$ Emotional Stability, computed as 6 $-$ Neuroticism.
In each panel, black solid lines indicate mean scores for the Unspecified expressive-style condition, whereas red and blue dashed lines indicate mean scores for the Emotional and Rational conditions, respectively.
The axis corresponding to the specified Big Five trait is highlighted in green.}
\vspace{-3mm}
  \includegraphics[width=0.5\linewidth]{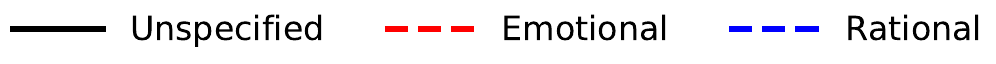} \\[1mm]
  \newcommand{\rowlabel}[1]{\raisebox{0.085\linewidth}{\makebox[28mm][l]{#1}}}
  \begin{tabular}{cccc}
  \hline

  % \diagbox[width=38mm,innerwidth=37mm]{\small Personality\\\small Conditions}{\hspace{-6mm}\small Roles} & Chat & Salesperson & Customer \\
  \diagbox[width=38mm,innerwidth=37mm]{\small Personality}{\hspace{-6mm}\small Role} & \rule{0pt}{0mm}Chat & Salesperson & Customer \\
  \hline
    \noalign{\vskip 2.5mm}
    \rowlabel{Unspecified} & 
    \vspace{0.5mm}
    \includegraphics[width=0.1805\linewidth]{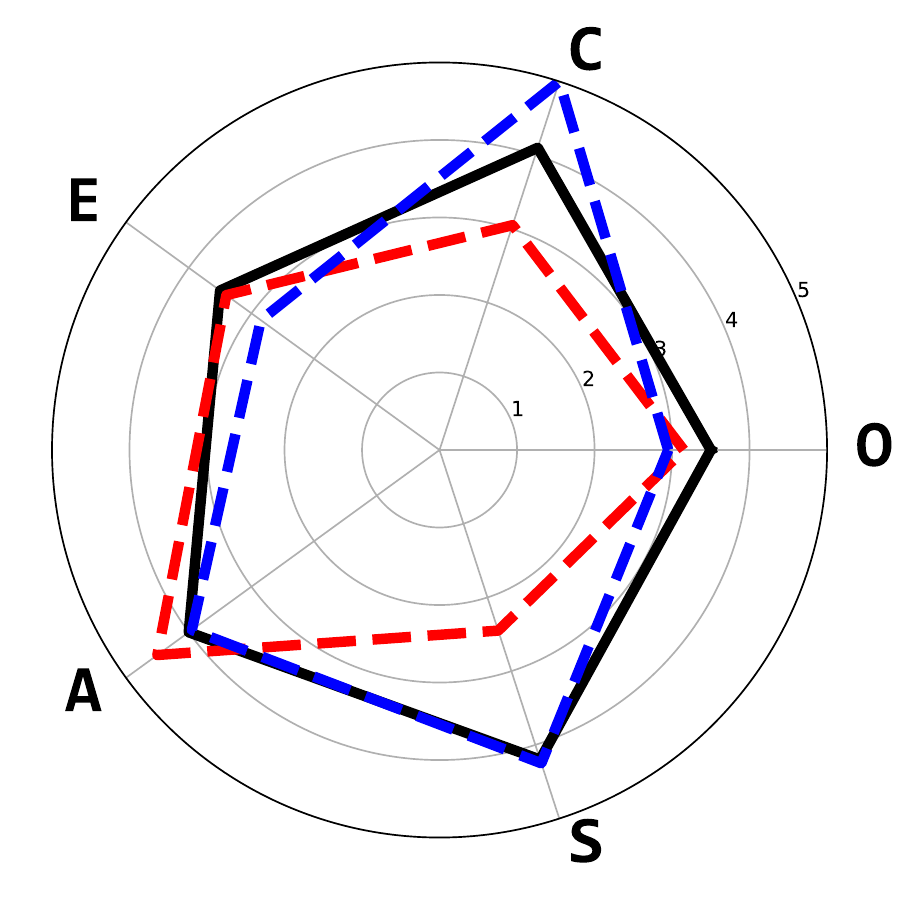} & 
    \vspace{0.5mm}
    \includegraphics[width=0.1805\linewidth]{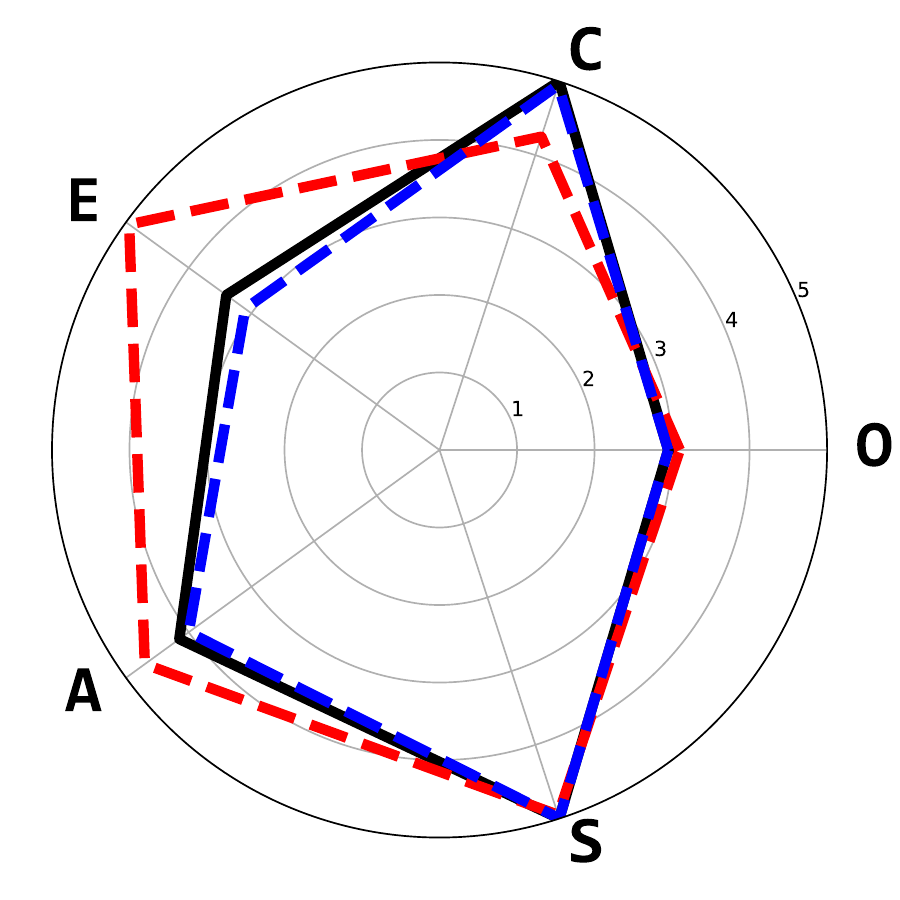} & 
    \vspace{0.5mm}
    \includegraphics[width=0.1805\linewidth]{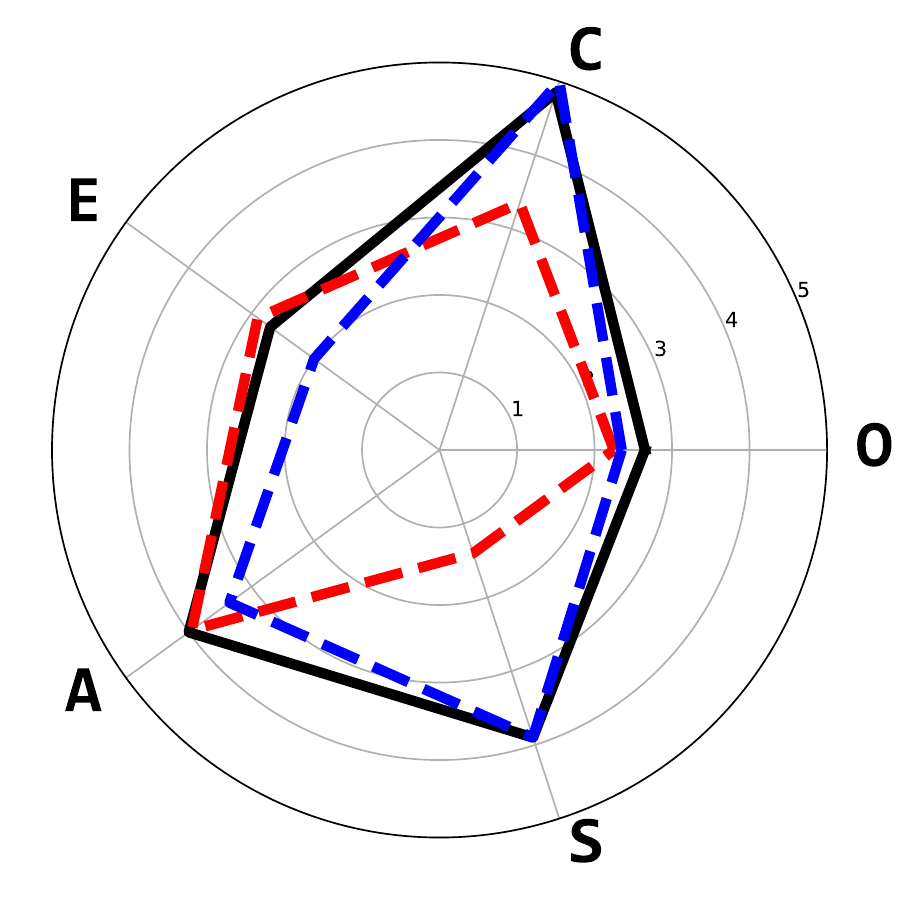} \\ 
    \hline
    \noalign{\vskip 2.5mm}
    \rowlabel{Openness} & 
    \vspace{0.5mm}
    \includegraphics[width=0.1805\linewidth]{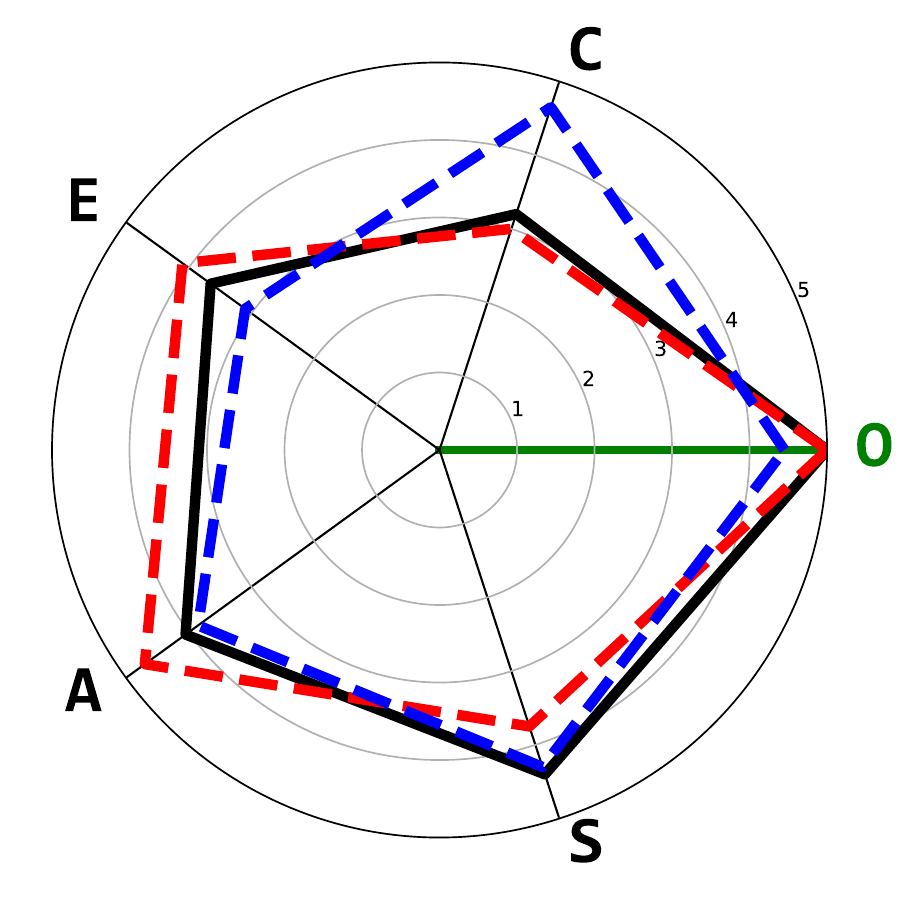} & 
    \vspace{0.5mm}
    \includegraphics[width=0.1805\linewidth]{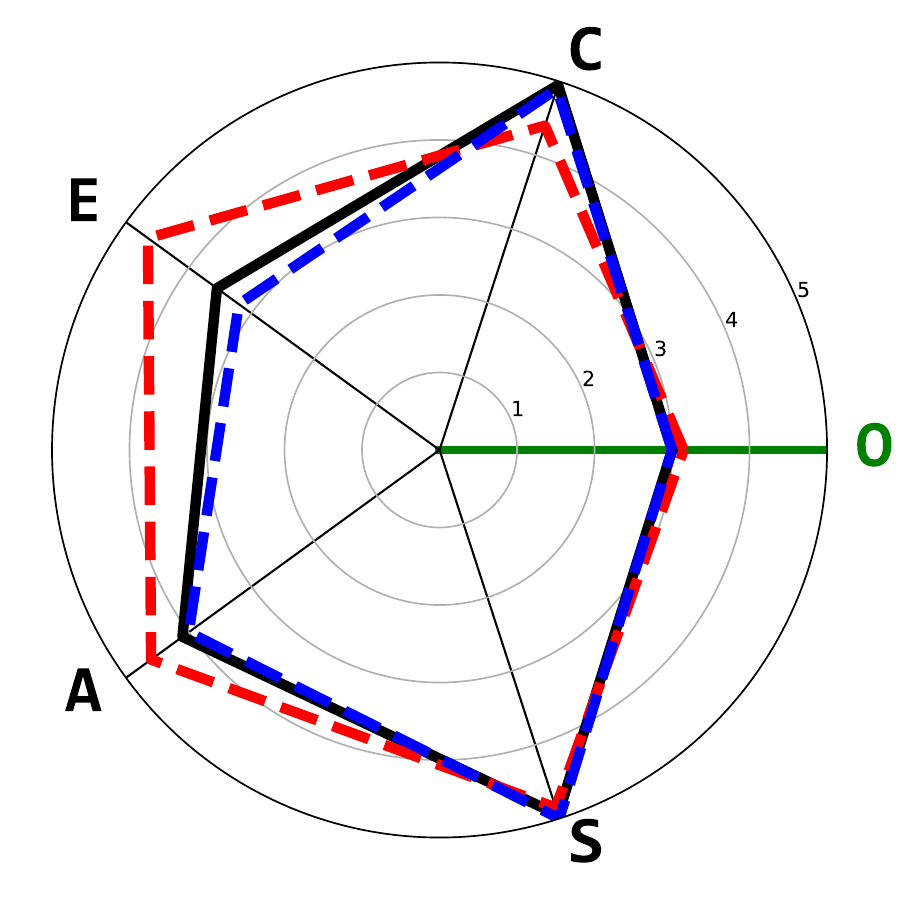} & 
    \vspace{0.5mm}
    \includegraphics[width=0.1805\linewidth]{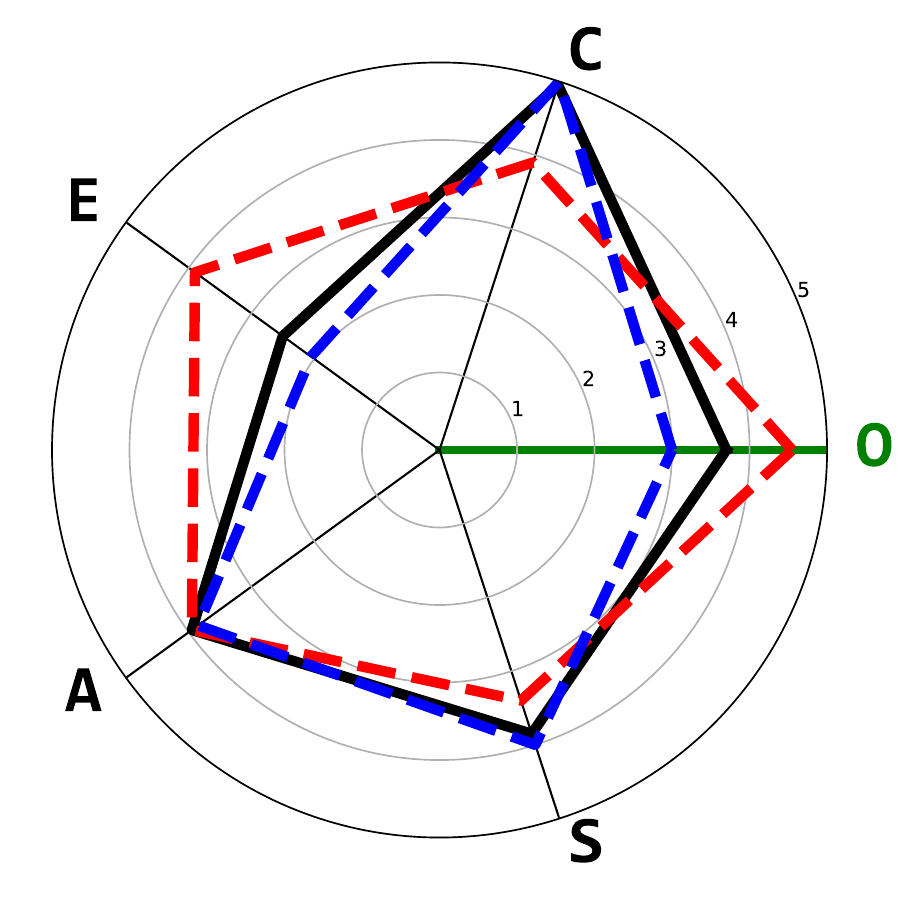} \\ 
    \hline
    \noalign{\vskip 2.5mm}
    \rowlabel{Conscientiousness} & 
    \vspace{0.5mm}
    \includegraphics[width=0.1805\linewidth]{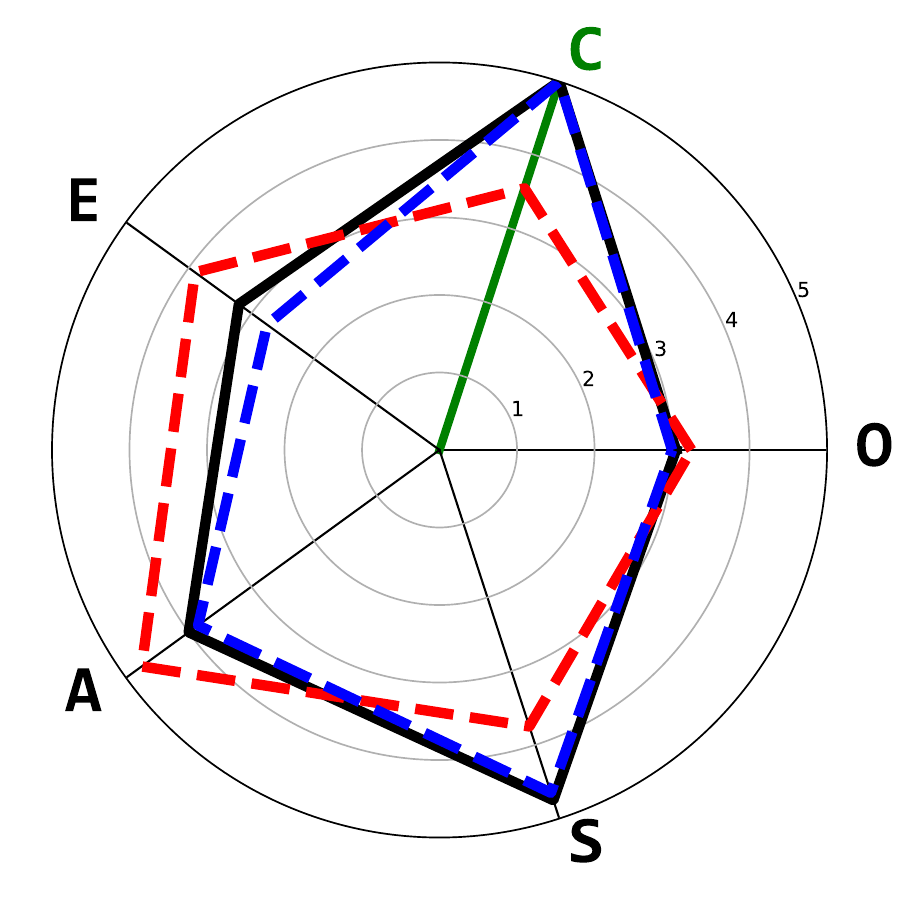} & 
    \vspace{0.5mm}
    \includegraphics[width=0.1805\linewidth]{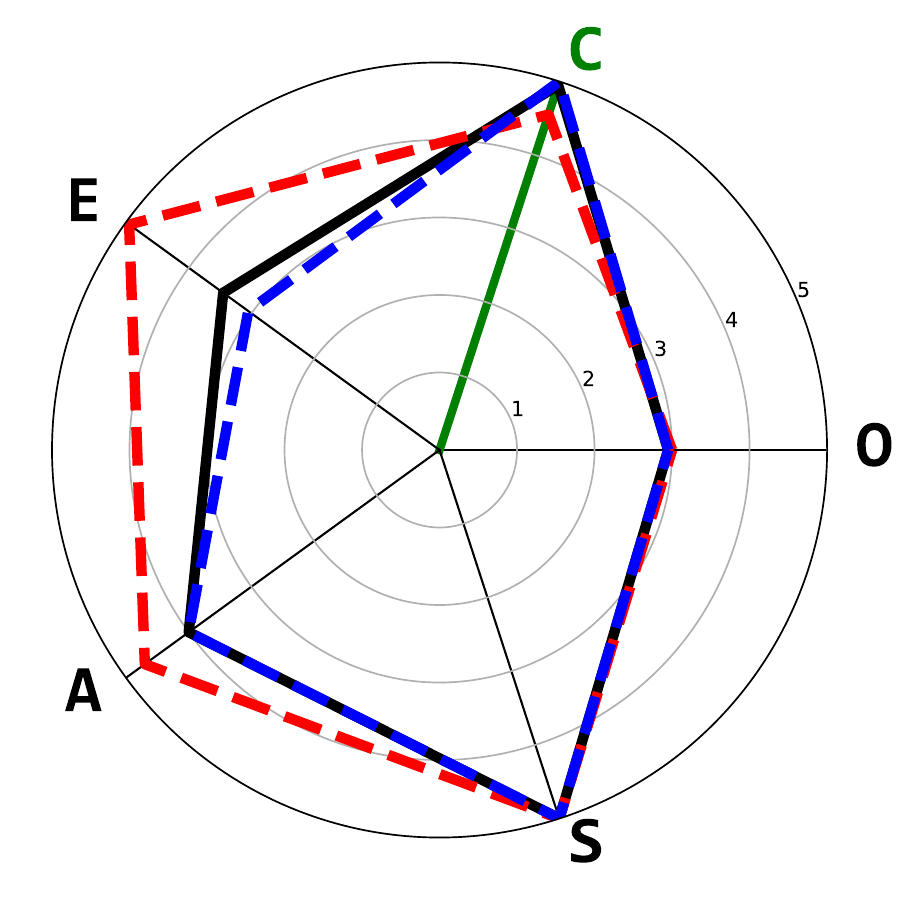} & 
    \vspace{0.5mm}
    \includegraphics[width=0.1805\linewidth]{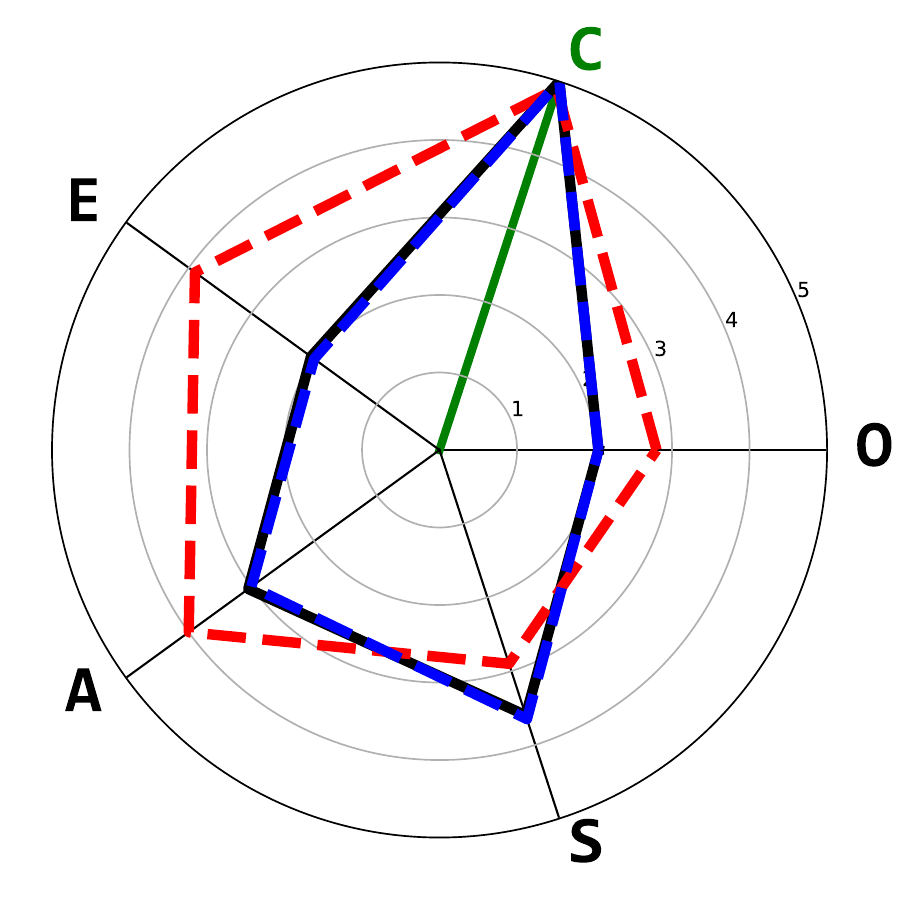} \\ 
    \hline
    \noalign{\vskip 2.5mm}
    \rowlabel{Extraversion} & 
    \vspace{0.5mm}
    \includegraphics[width=0.1805\linewidth]{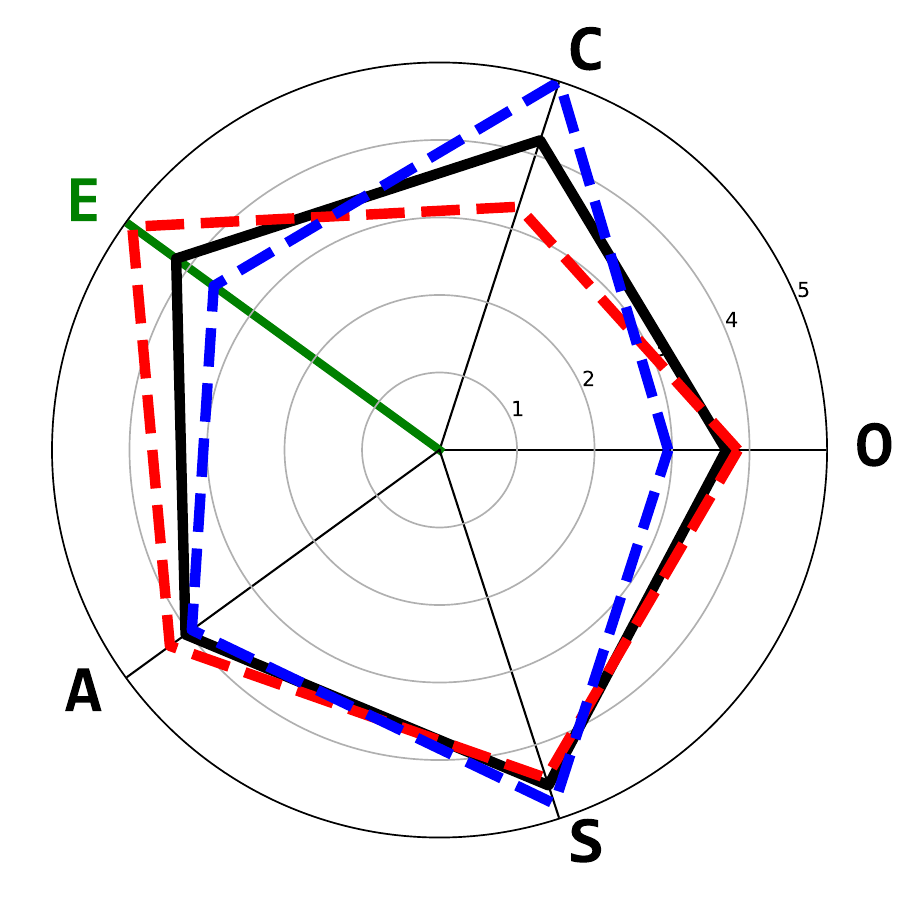} & 
    \vspace{0.5mm}
    \includegraphics[width=0.1805\linewidth]{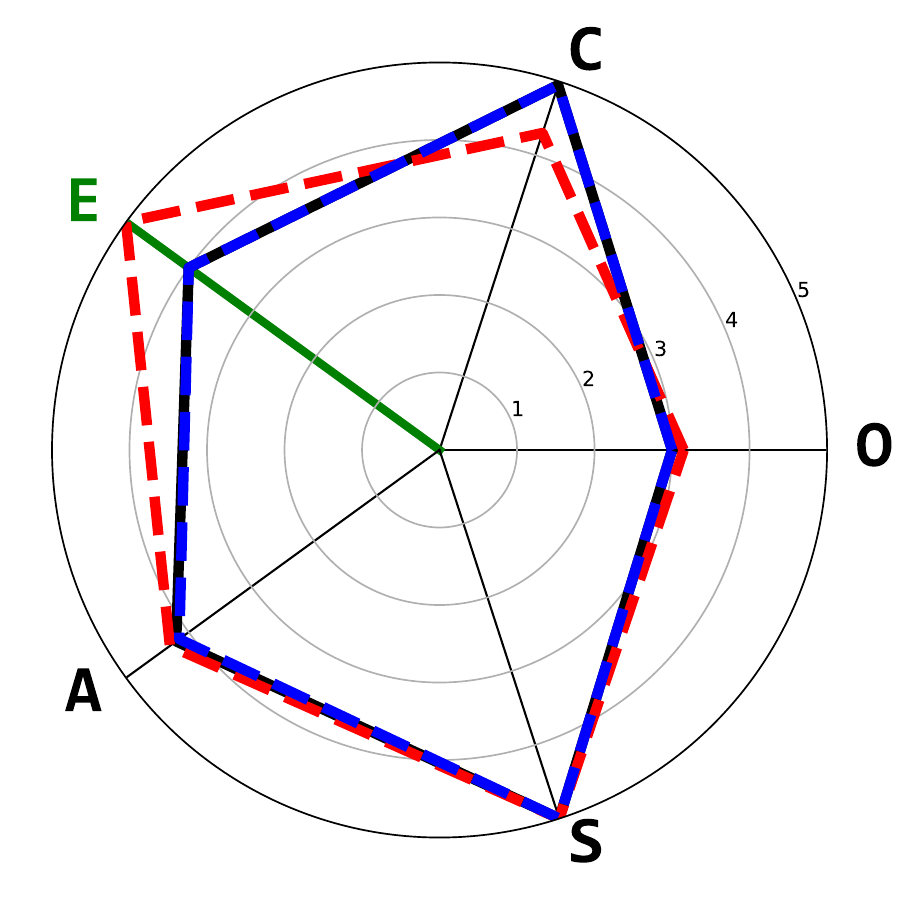} & 
    \vspace{0.5mm}
    \includegraphics[width=0.1805\linewidth]{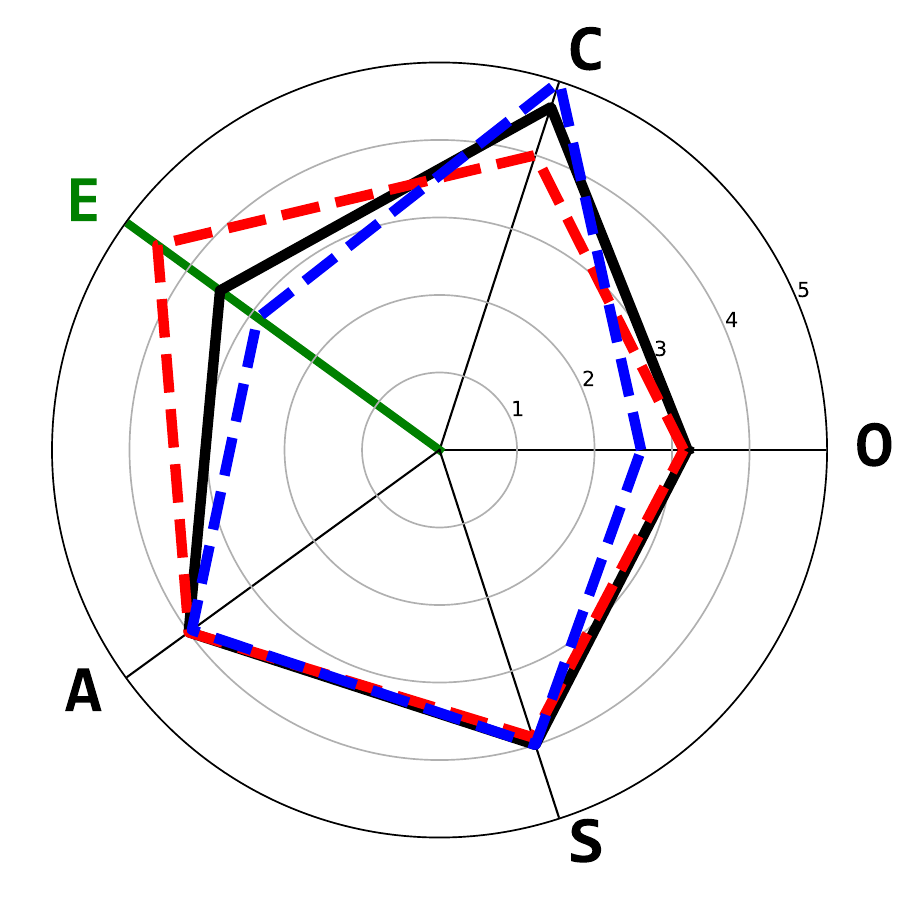} \\ 
    \hline
    \noalign{\vskip 2.5mm}
    \rowlabel{Agreeableness} &
    \vspace{0.5mm}
    \includegraphics[width=0.1805\linewidth]{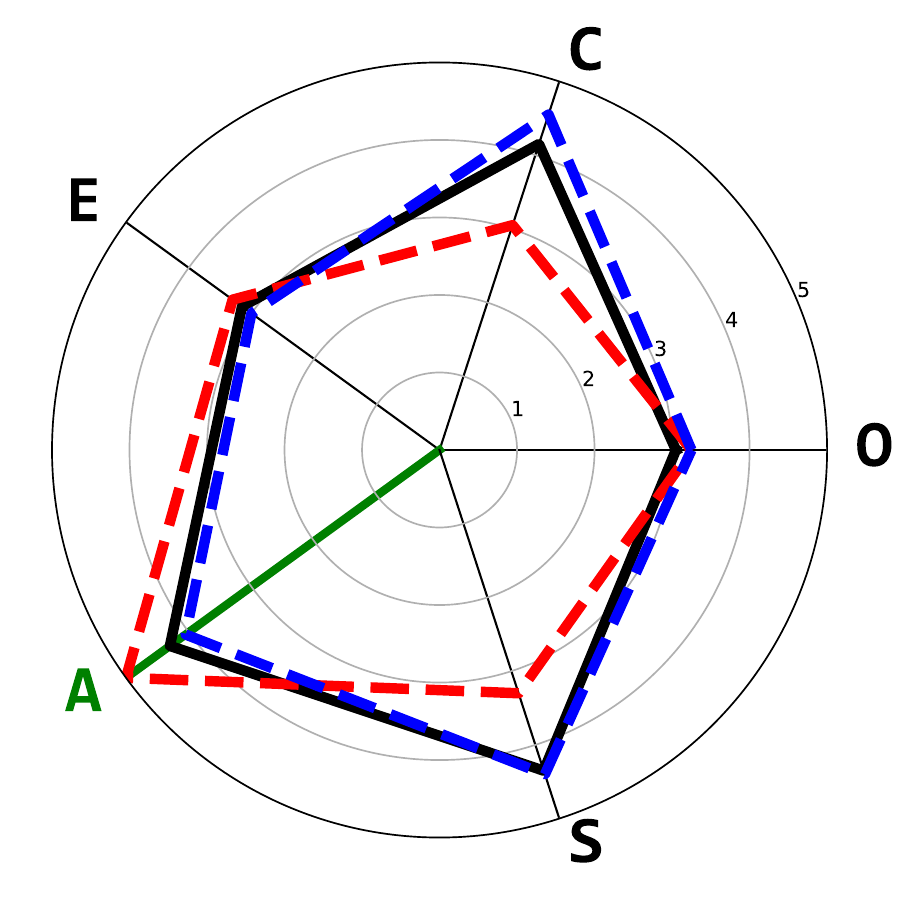} & 
    \vspace{0.5mm}
    \includegraphics[width=0.1805\linewidth]{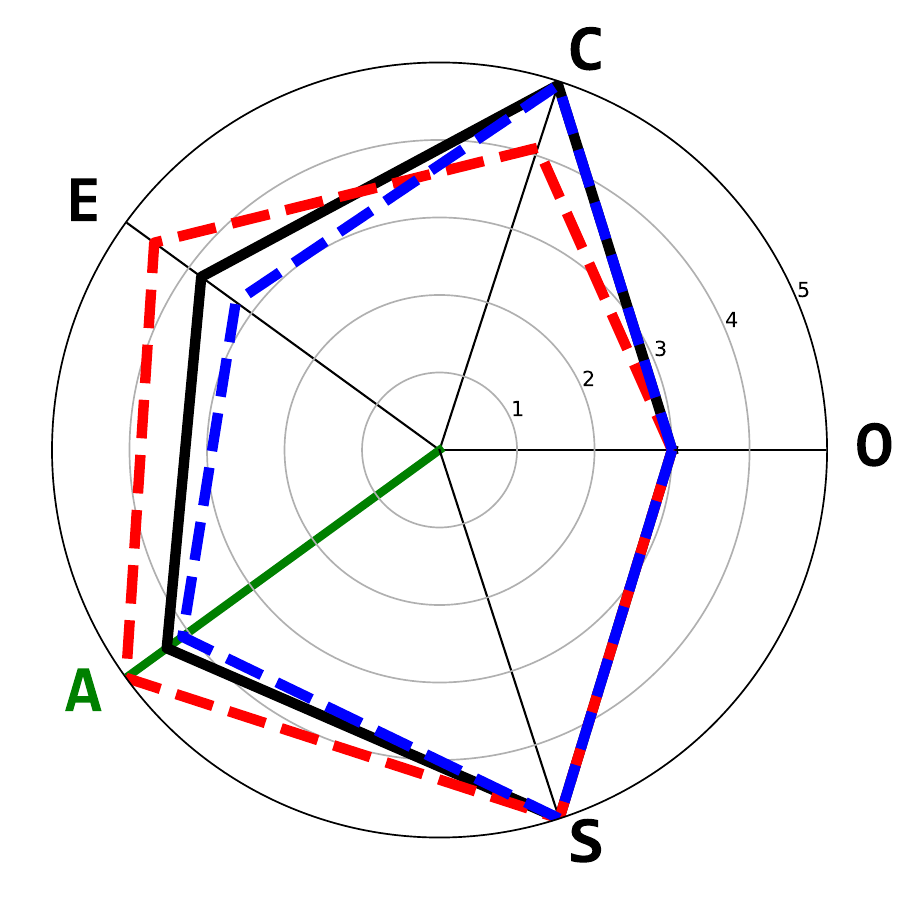} & 
    \vspace{0.5mm}
    \includegraphics[width=0.1805\linewidth]{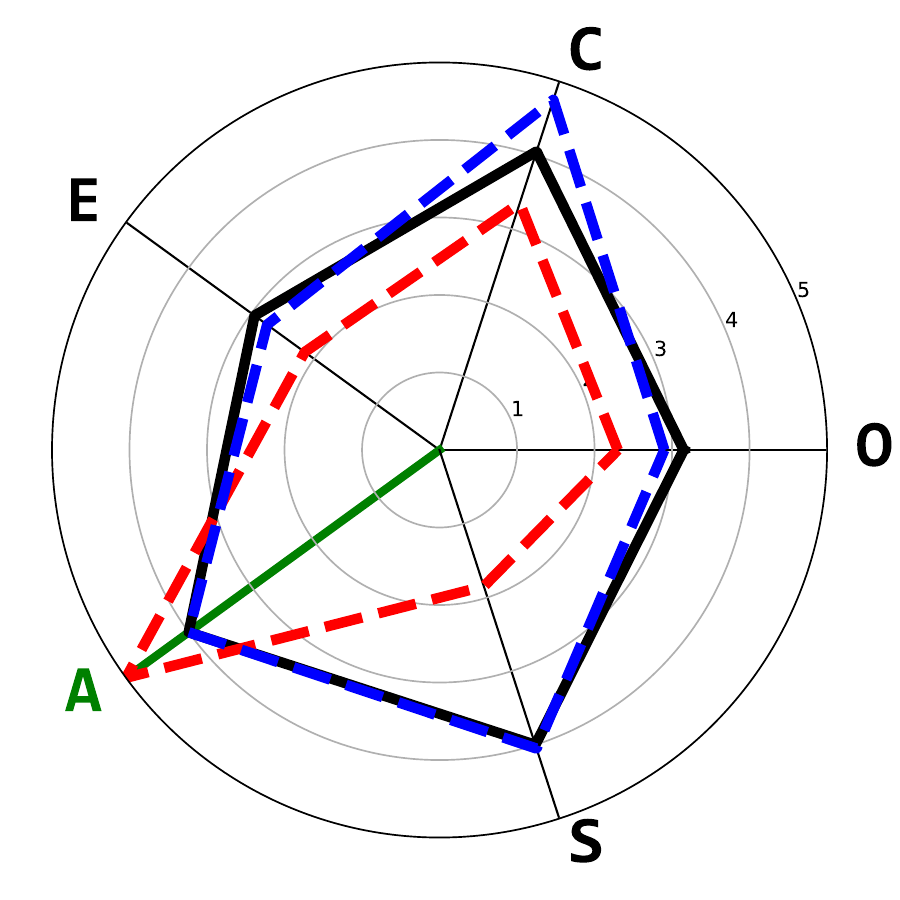} \\ 
    \hline
    \noalign{\vskip 2.5mm}
    \rowlabel{Neuroticism} & 
    \vspace{0.5mm}
    \includegraphics[width=0.1805\linewidth]{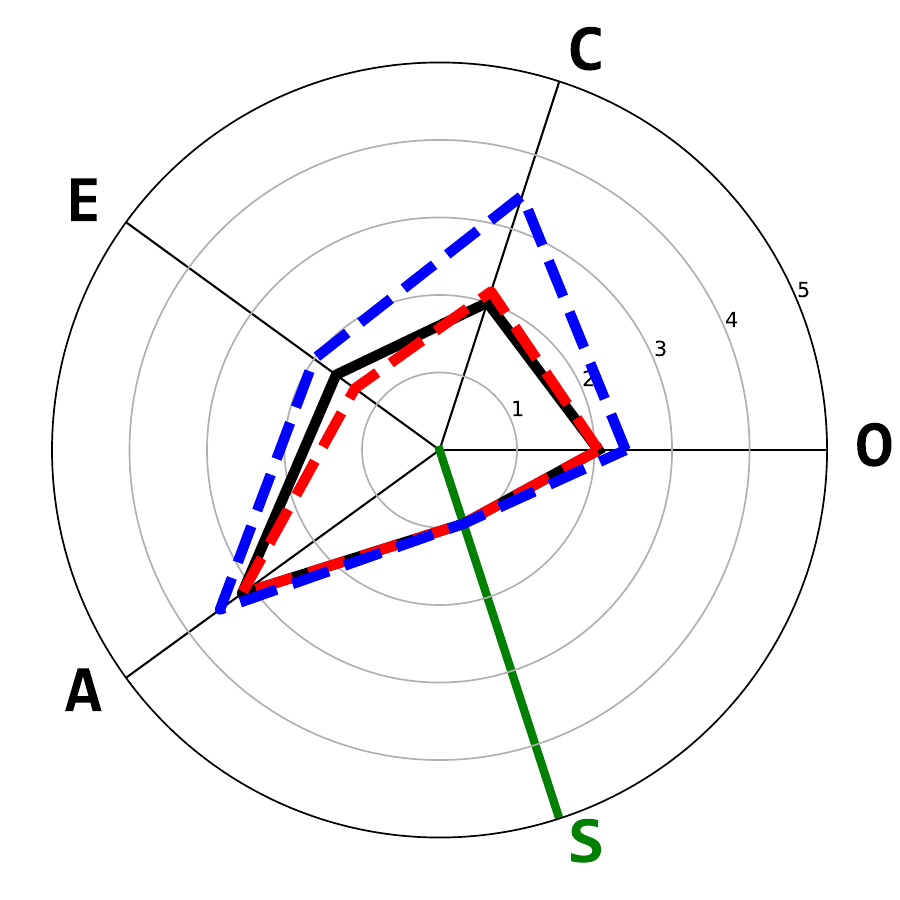} & 
    \vspace{0.5mm}
    \includegraphics[width=0.1805\linewidth]{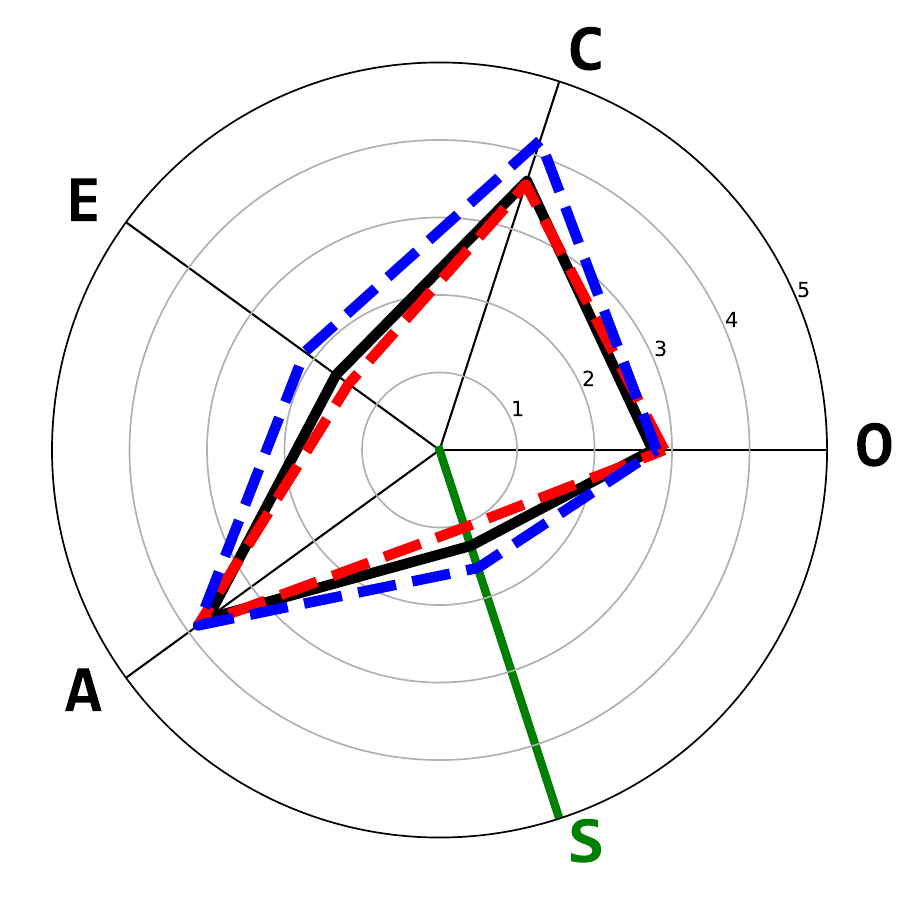} & 
    \vspace{0.5mm}
    \includegraphics[width=0.1805\linewidth]{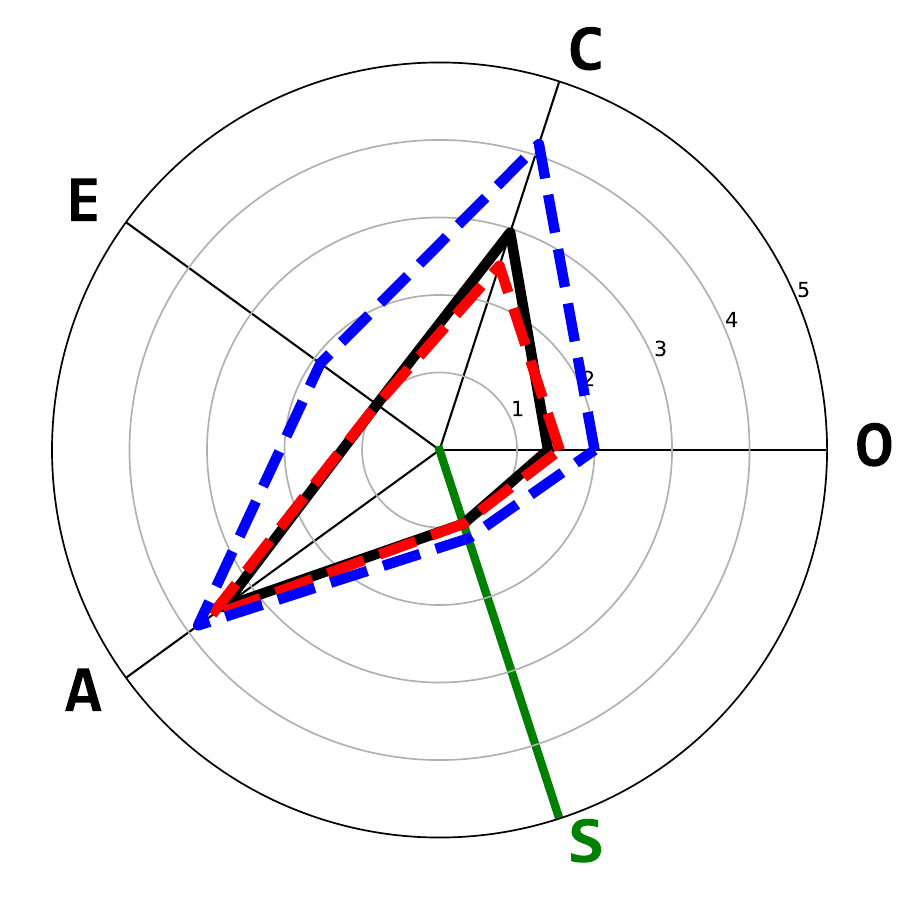} \\ 
    \hline
  \end{tabular}
  \label{tab:radar-personality-role-expressive-style}

% \vspace{-10mm}

\end{table}

% \begin{itemize}
%     \item[{\bf --}] 
%         Overall, Expressive Style either facilitated or suppressed the manifestation of the trait corresponding to the personality condition, and this effect was role-dependent.
%     \item[{\bf --}] 
%         Across Roles, the Emotional expressive-style condition facilitated the manifestation of the specified traits under the Extraversion and Agreeableness personality conditions.
%     \item[{\bf --}] 
%         In the Chat role, under the Conscientiousness personality condition, the Conscientiousness score was lower under the Emotional expressive-style condition than under the Unspecified expressive-style condition, whereas the score did not change in the Salesperson and Customer roles.
%     \item[{\bf --}] 
%         In the Customer role, under the Openness personality condition, the Openness score was higher under the Emotional expressive-style condition than under the Unspecified expressive-style condition, whereas the scores were the same in the Chat and Salesperson roles.
% \end{itemize}

\subsubsection*{Two-Dimensional Configuration of Mean Big Five Trait Scores}

\begin{figure}[t]
  \centering
  \includegraphics[width=0.9\linewidth]{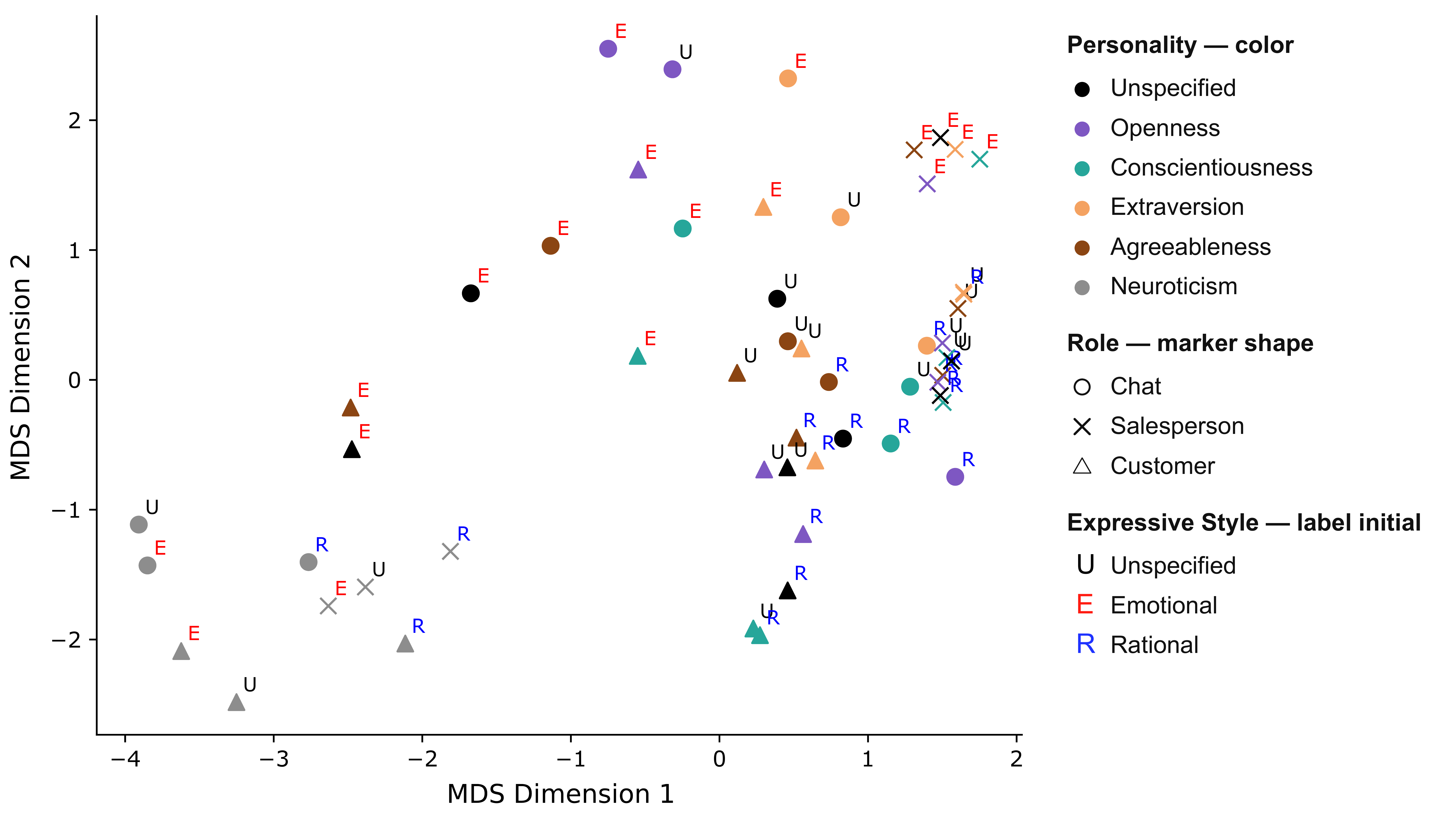}
  \caption{Two-dimensional configuration of the pentagonal mean Big Five score polygons in the radar charts for joint effects of Personality, Role, and Expressive Style.}
  \label{fig:Personality_Role_Expressive_style_MDS}
\end{figure}

Figure~\ref{fig:Personality_Role_Expressive_style_MDS} shows a two-dimensional configuration of the pentagonal mean-score polygons in the radar charts, obtained by multidimensional scaling (MDS). The MDS solution yielded a normalized stress value of 0.086, indicating a reasonably good two-dimensional representation of the original dissimilarities.

This exploratory visualization provides an overview of how the overall Big Five profiles varied across the full combination of experimental factors. The spread of points within each Personality condition indicates that Role and Expressive Style introduced substantial variation even when the same Personality condition was specified. In other words, the profiles were not organized solely by Personality condition. Instead, the relative positions of the points suggest that dialogue-level factors, especially Role and Expressive Style, contributed strongly to the overall configuration of expressed personality profiles. Because the MDS analysis is used here as an exploratory visualization, the primary statistical evidence is provided by the ANOVA results below.

\subsubsection*{Analysis of Variance}

Table~\ref{tab:three_way_anova} presents the results of three-way ANOVAs examining the effects of Personality, Role, and Expressive Style on each expressed Big Five scale. We conducted five separate analyses, one for each Big Five trait dimension as the dependent variable. In each analysis, Personality had two levels, Unspecified and the corresponding dimension-specific condition; Role had three levels, Chat, Salesperson, and Customer; and Expressive Style had three levels, Unspecified, Emotional, and Rational.

The ANOVA results show that expressed Big Five traits were systematically affected by all three factors, but the dominant source of variation differed across trait dimensions. For Openness, Personality showed the largest effect ($\omega^2 = .3053$), followed by Role ($\omega^2 = .2471$), and the Personality $\times$ Role interaction was also substantial ($\omega^2 = .1410$). This indicates that Openness expression was strongly influenced by explicit Openness specification, but also depended on whether the agent was placed in an unconstrained chat setting or in a task-oriented retail interaction.

For Conscientiousness and Agreeableness, Expressive Style showed the largest effects ($\omega^2 = .4078$ and $\omega^2 = .4024$, respectively). These results indicate that these two trait dimensions were particularly sensitive to whether the agent responded in an Emotional or Rational style. Extraversion was influenced by all three factors, with large effects of Expressive Style ($\omega^2 = .2744$), Personality ($\omega^2 = .2120$), and Role ($\omega^2 = .1910$), suggesting that Extraversion expression was shaped jointly by trait specification, social role, and expressive orientation. Neuroticism was dominated by Personality ($\omega^2 = .6845$), which was the largest effect observed in this analysis.

Although all three-way interactions were statistically significant, their effect sizes were modest compared with the dominant main effects and major two-way interactions. Thus, the key finding is not merely that Personality, Role, and Expressive Style interact, but that their relative contributions differ substantially across Big Five dimensions. Overall, these results support an interactionist interpretation of LLM personality expression: expressed traits in generated dialogue are shaped not only by explicit personality prompts, but also by the role and expressive style in which the model responds.

\begin{table}[t]
\centering
\caption{Three-way ANOVA results for the effects of Personality, Role, and Expressive Style on
each expressed Big Five scale}
\label{tab:three_way_anova}
\begin{tabular}{llrrrr}
\hline
Dependent Variable & Factor/Interaction & df & F & p & $\omega^2$ \\
\hline
\multirow{7}{*}{Openness}
 & Personality & 1, 342 & 552.603 & $< .001$ & 0.3053 \\
 & Role & 2, 342 & 224.231 & $< .001$ & 0.2471 \\
 & Expressive Style & 2, 342 & 36.908 & $< .001$ & 0.0397 \\
 & Personality $\times$ Role & 2, 342 & 128.378 & $< .001$ & 0.1410 \\
 & Personality $\times$ Expressive Style & 2, 342 & 22.923 & $< .001$ & 0.0243 \\
 & Role $\times$ Expressive Style & 4, 342 & 9.074 & $< .001$ & 0.0179 \\
 & Personality $\times$ Role $\times$ Expressive Style & 4, 342 & 12.536 & $< .001$ & 0.0255 \\
\hline
\multirow{7}{*}{Conscientiousness}
 & Personality & 1, 342 & 101.685 & $< .001$ & 0.0685 \\
 & Role & 2, 342 & 66.500 & $< .001$ & 0.0891 \\
 & Expressive Style & 2, 342 & 300.635 & $< .001$ & 0.4078 \\
 & Personality $\times$ Role & 2, 342 & 14.338 & $< .001$ & 0.0182 \\
 & Personality $\times$ Expressive Style & 2, 342 & 35.251 & $< .001$ & 0.0466 \\
 & Role $\times$ Expressive Style & 4, 342 & 27.214 & $< .001$ & 0.0714 \\
 & Personality $\times$ Role $\times$ Expressive Style & 4, 342 & 20.638 & $< .001$ & 0.0535 \\
\hline
\multirow{7}{*}{Extraversion}
 & Personality & 1, 342 & 314.547 & $< .001$ & 0.2120 \\
 & Role & 2, 342 & 142.190 & $< .001$ & 0.1910 \\
 & Expressive Style & 2, 342 & 203.920 & $< .001$ & 0.2744 \\
 & Personality $\times$ Role & 2, 342 & 13.300 & $< .001$ & 0.0166 \\
 & Personality $\times$ Expressive Style & 2, 342 & 4.304 & 0.014 & 0.0045 \\
 & Role $\times$ Expressive Style & 4, 342 & 12.195 & $< .001$ & 0.0303 \\
 & Personality $\times$ Role $\times$ Expressive Style & 4, 342 & 11.264 & $< .001$ & 0.0278 \\
\hline
\multirow{7}{*}{Agreeableness}
 & Personality & 1, 342 & 131.017 & $< .001$ & 0.1169 \\
 & Role & 2, 342 & 41.583 & $< .001$ & 0.0730 \\
 & Expressive Style & 2, 342 & 224.769 & $< .001$ & 0.4024 \\
 & Personality $\times$ Role & 2, 342 & 12.411 & $< .001$ & 0.0205 \\
 & Personality $\times$ Expressive Style & 2, 342 & 18.840 & $< .001$ & 0.0321 \\
 & Role $\times$ Expressive Style & 4, 342 & 1.088 & 0.363 & 0.0003 \\
 & Personality $\times$ Role $\times$ Expressive Style & 4, 342 & 9.628 & $< .001$ & 0.0310 \\
\hline
\multirow{7}{*}{Neuroticism}
 & Personality & 1, 342 & 3444.500 & $< .001$ & 0.6845 \\
 & Role & 2, 342 & 193.056 & $< .001$ & 0.0764 \\
 & Expressive Style & 2, 342 & 130.597 & $< .001$ & 0.0515 \\
 & Personality $\times$ Role & 2, 342 & 106.167 & $< .001$ & 0.0418 \\
 & Personality $\times$ Expressive Style & 2, 342 & 82.125 & $< .001$ & 0.0323 \\
 & Role $\times$ Expressive Style & 4, 342 & 19.806 & $< .001$ & 0.0150 \\
 & Personality $\times$ Role $\times$ Expressive Style & 4, 342 & 35.042 & $< .001$ & 0.0271 \\
\hline
\end{tabular}
\end{table}

%%%%%%%%%%%%%%%%%%%%%%%%%%%%%%%%%%%%%%%%%%%%%%%%%%%%%%%%%%%%%
\subsection{Comparison of English and Japanese Dialogues}

To examine cross-linguistic differences, we compared the mean Big Five trait scores obtained from English and Japanese dialogues under corresponding experimental conditions. Table~\ref{tab:rmse_between_enjp} reports the RMSE values between English and Japanese scores for each Big Five trait, using Gemini 2.5 Flash as the evaluator in both languages.

Overall, the RMSE values indicate that cross-linguistic differences were generally limited. However, relatively large differences were observed under a small number of specific Personality--Role--Expressive Style combinations. The two conditions with the largest differences were the Unspecified--Chat--Emotional condition and the Neuroticism--Salesperson--Rational condition.

% For the cross-linguistic comparison, we compared the mean trait scores obtained from English and Japanese dialogues under corresponding experimental conditions.
% Table~\ref{tab:rmse_between_enjp} shows language-dependent differences by comparing English and Japanese dialogues, using Gemini 2.5 Flash (gemini-2.5-flash) as the evaluator in both cases.
% The RMSE values indicate that language-dependent differences were generally small.
% Overall, cross-linguistic differences between English and Japanese dialogues were small.
% Relatively large differences were observed only for a limited number of specific Personality--Role--Expressive Style combinations.
% The two conditions with the largest differences are the Unspecified--Chat--Emotional condition and the Neuroticism--Salesperson--Rational condition.

\paragraph{Unspecified--Chat--Emotional condition}

Figure~\ref{tsuika1} compares the Big Five trait scores for English and Japanese dialogues in the Unspecified--Chat--Emotional condition. In the English dialogues, Emotional Stability decreased substantially, indicating an increase in perceived Neuroticism. In contrast, this decrease was less pronounced in the Japanese dialogues. This suggests that emotionally expressive chat responses were interpreted as more neurotic in English than in Japanese under otherwise identical prompt conditions.
To illustrate how expressive style affected the generated dialogue, Table~\ref{table:dialogue_example_1} presents an example of an English dialogue in the Unspecified--Chat--Emotional condition, whereas Table~\ref{table:dialogue_example_2} presents a corresponding example in the Unspecified--Chat--Rational condition. These examples provide qualitative context for the quantitative difference observed in the Unspecified--Chat condition.

\begin{table*}[t]
\centering
\caption{RMSE between Big Five trait scores of English- and Japanese-language dialogues, using Gemini
2.5 Flash as the evaluator.}
\label{tab:rmse_between_enjp}
\begin{tabular}{lcc}
\hline
Trait & RMSE\\
\hline
Openness & 0.35 \\
Conscientiousness & 0.42 \\
Extraversion & 0.40 \\
Agreeableness & 0.30 \\
Neuroticism & 0.50 \\
\hline
\end{tabular}
\end{table*}

\begin{figure*}[t]
\vspace{4mm}
\input{figure_set_eng_jpn}
\end{figure*}

\paragraph{Neuroticism--Salesperson--Rational condition}

Figure~\ref{tsuika2} compares the Big Five trait scores for English and Japanese dialogues in the Neuroticism--Salesperson--Rational condition. When the Neuroticism personality condition was specified, Emotional Stability was low in both languages overall. However, under the Salesperson role with the Rational expressive style, Emotional Stability remained relatively higher in Japanese than in English. This suggests that the expression of Neuroticism was attenuated under a specific combination of conditions: Japanese language, Salesperson role, and Rational expressive style.

Taken together, these results suggest that language did not uniformly alter personality expression. Rather, cross-linguistic differences appeared in a condition-dependent manner, emerging only under specific combinations of Personality, Role, and Expressive Style.

\section{Discussion}
\label{sec:discussion}

\paragraph{Main Findings.}
Our analysis yielded three main findings. First, expressed Big Five traits in LLM-generated dialogue were not determined solely by explicit personality-trait specifications. Role and Expressive Style also systematically shaped perceived trait expression, even when no Big Five trait was explicitly assigned. Second, the relative importance of Personality, Role, and Expressive Style differed substantially across Big Five dimensions. Personality specification was the dominant source of variation for Openness and Neuroticism, whereas Expressive Style had the largest effects on Conscientiousness, Extraversion, and Agreeableness. Third, the effect of personality specification depended on the dialogue context. In particular, the Personality $\times$ Role and Personality $\times$ Expressive Style interactions showed that the same personality prompt could be expressed differently depending on the assigned role and expressive style. Although the three-way interactions were statistically significant, their effect sizes were modest compared with the dominant main effects and two-way interactions. Thus, the central finding is not simply that all three factors interact, but that the structure of expressed personality differs across trait dimensions.

\paragraph{Trait-specific Patterns of Personality Expression.}

The observed effects suggest that each Big Five dimension is sensitive to different aspects of the dialogue setting. Openness was strongly influenced by both Personality and Role: explicit Openness specification increased expressed Openness, but the magnitude of this expression also depended on whether the agent was placed in an unconstrained chat setting or in a more task-oriented retail interaction. Neuroticism, by contrast, was dominated by Personality specification, suggesting that prompts specifying high Neuroticism produced a strong and relatively stable effect across roles and expressive styles.
Conscientiousness and Agreeableness were especially sensitive to Expressive Style. Rational expression tended to increase perceived Conscientiousness, whereas Emotional expression tended to increase perceived Agreeableness. Extraversion was influenced by all three factors, suggesting that its expression is shaped jointly by explicit trait specification, social role, and expressive orientation. 
These trait-specific patterns indicate that personality control in LLMs cannot be understood as a uniform effect of trait prompting; rather, each trait dimension appears to interact with contextual factors in a distinct way.

\paragraph{Personality in Social Situations.}

These findings support an interactionist view of LLM personality expression. In this view, expressed traits are not direct readouts of internal personality prompts, but context-dependent outcomes shaped by the interplay between trait specifications and social conditions. The present results suggest that role and expressive style can amplify, suppress, or redirect the expression of a specified personality trait. Importantly, the form of this contextual modulation depends on the target trait dimension. This trait-specific pattern is consistent with interactionist accounts of personality, which emphasize that behavioral expression depends on both dispositional tendencies and situational constraints.

\paragraph{Implications for Controllable Dialogue Agents.}

The results have practical implications for the design of LLM-based dialogue agents. If personality control is implemented only by specifying Big Five traits in prompts, the resulting utterances may not consistently express the intended traits across different roles or expressive styles. For example, a trait specification that is effective in an unconstrained chat setting may be weakened or transformed in a task-oriented role. Similarly, an expressive-style instruction may shift perceived personality even in the absence of explicit trait specification.
Therefore, controllable agent design should treat personality, role, and expressive style as jointly operating control factors rather than as independent prompt components. In particular, personality-controlled agents should be evaluated under the specific social roles and expressive conditions in which they are intended to operate.

\paragraph{Cross-linguistic Differences.}
The comparison between English and Japanese dialogues showed generally limited but condition-dependent cross-linguistic differences. This suggests that language does not uniformly determine personality expression in LLM agents. Rather, language may modulate expressed traits in interaction with role and expressive style under specific conditions. For example, differences were observed in emotionally expressive chat and in rational salesperson interactions under high Neuroticism. These findings indicate that multilingual evaluation is important, but they also suggest that cross-linguistic effects should be interpreted cautiously and in relation to specific social and expressive contexts.

\paragraph{From Prompt-based to Latent-space Control.}
Although this study focused on prompt-based control, the interactionist perspective has implications for other approaches to controlling LLM behavior, including latent-space steering. Recent work has explored methods for controlling personality or style through internal representations~(\citealp{frising2025linear}; \citealp{wu2025ai}; \citealp{wu2026humanlm}). However, if personality expression depends on role and expressive style, then steering a single personality dimension in isolation may not be sufficient to produce stable behavior across contexts. Future work should therefore examine whether the joint effects observed here are reflected in the latent representations of LLMs and whether such representations can support more robust, context-sensitive personality control.

\paragraph{Cognitive Perspectives.}
Our findings do not imply that LLMs possess personality in the psychological sense. Rather, they show that LLM-generated utterances can produce personality-like impressions that vary systematically across dialogue contexts. Even when the same personality specification was used, perceived Big Five traits changed depending on the assigned role and expressive style. This pattern is analogous to the situation-dependent expression of personality traits in humans, where the same individual may express different aspects of personality across different social roles. A related interaction between Big Five traits and social roles in humans has also been reported by \citet{sheldon1997trait}.
Thus, interactionist personality theory provides a useful framework for analyzing how socially situated prompt conditions shape the personality-like impressions conveyed by LLM-generated dialogue. More broadly, these limited parallels suggest that comparative cognition may offer a productive perspective for studying socially situated behavior in computational agents~\citep{voudouris2025bringing}.

%%%%%%%%%%%%%%%%%%%%%%%%%%%%%%%%%%%
\section{Related Work}
\label{sec:related_work}

Before the advent of LLMs, work on personality and persona control in dialogue systems followed several representative paradigms. 
Early approaches controlled personality-related expression through overgenerate-and-rank frameworks based on statistical models of personality and alignment~\citep{isard2006individuality}, or through explicit generation parameters derived from psycholinguistic correlates of personality, such as verbosity, polarity, self-references, hedges, and lexical choice~\citep{mairesse2007personage}. 
Later neural approaches shifted the focus toward persona-grounded dialogue, where responses are conditioned on speaker profile sentences to improve persona consistency~\citep{zhang2018personalizing}. 
These approaches differ in whether they control personality or persona through statistical ranking, explicit linguistic parameters, or profile-based conditioning.

The advent of LLMs has substantially changed this design space. 
Rather than relying primarily on hand-engineered generation parameters, task-specific rankers, or supervised persona-conditioned dialogue models, LLM-based systems can be steered through natural-language instructions, role descriptions, in-context demonstrations, retrieved memories, or preference-based optimization. 
This allows personality to be specified at a higher level of abstraction, while relying on the pretrained model to realize the corresponding linguistic, pragmatic, and interactional patterns. 
However, this flexibility also reduces transparency and controllability: it is often unclear which linguistic cues give rise to a perceived personality, whether the model will maintain the intended persona across turns, and how personality specifications interact with task goals, safety constraints, and other prompt attributes.

Recent work on LLM-based personality and persona control has therefore examined both how desired personalities can be induced and how their expression should be evaluated. 
Existing approaches span prompting-based methods, training-based methods such as supervised fine-tuning and preference optimization, and representation-level steering methods. 
Since our study focuses on prompting as a lightweight and widely used control mechanism, we primarily review prompting-based work. 
Within prompting-based research, prior studies have developed along several parallel strands, as summarized in Table~\ref{tab:related_work_comparison}.

One line of work investigates whether LLMs can be specified and evaluated along human personality dimensions. 
\citet{jiang-etal-2024-personallm} analyzed stories generated by LLMs prompted with Big Five traits, using human evaluation, LLM-based evaluation, and Linguistic Inquiry and Word Count (LIWC)~\citep{pennebaker2001liwc}. 
\citet{nature_machine_intelligence} examined LLM behavior under demographic, contextual, and linguistic variation using psychometric instruments such as the Big Five Inventory (BFI)~\citep{john1991bfi} and the International Personality Item Pool--NEO (IPIP-NEO)~\citep{goldberg2006ipip}.

Beyond explicit personality traits, another body of work shows that social specifications can shape LLM behavior. 
Persona and role assignments have been shown to influence response style, decision-making tendencies, and downstream task performance. 
\citet{chen2024persona} surveyed persona-based prompting and classified personas into demographic, character-based, and individualized categories. 
\citet{kong2024better} investigated role-play prompting as a zero-shot method for improving LLM reasoning, showing that task-relevant roles can outperform standard zero-shot prompting and, in many cases, Zero-Shot-CoT across diverse reasoning benchmarks. 
\citet{cao2026biased} showed that demographic personas assigned to LLM agents can systematically alter behavior and degrade performance across a range of agentic tasks.

Expressive style provides another source of variation that may interact with perceived personality. 
\citet{reif2022recipe} proposed a prompting-based method for transferring both standard styles, such as sentiment and formality, and non-standard styles, such as comic, melodramatic, metaphorical, and descriptive styles, as well as style specifications involving particular lexical choices.

Affective prompting further demonstrates that emotional cues can shape model outputs. 
\citet{li2023large} proposed EmotionPrompt, which augments standard prompts with psychology-inspired emotional stimuli, and showed that such prompts can improve LLM performance, truthfulness, and responsible response generation.

Although these studies show that personality traits, roles, styles, and affective cues can each shape LLM behavior, they typically examine these factors in isolation. 
In contrast, we adopt the interactionist perspective of \citet{fleeson2015whole}, which views personality expression as emerging from the interaction between internal dispositions and situational factors. 
From this perspective, prompt-specified Big Five traits are not sufficient explanations of observed behavior: their expression may be amplified, attenuated, or transformed by the agent's role, expressive style, and dialogue context. 
Building on this view, we examine how Big Five trait prompts manifest in LLM-generated dialogues across roles, styles, and languages.

\begin{table*}[t]
\caption{Control factors, targets, and evaluation methods for prior work and our approach.}
\label{tab:related_work_comparison}
\centering
\small
\begin{tabular}{p{2.4cm}p{2.8cm}p{2.2cm}p{6.8cm}}
\toprule
\textbf{Study} & \textbf{Control factor} & \textbf{Target} & \textbf{Evaluation} \\
\midrule
\citet{reif2022recipe}
& Standard and non-standard expressive styles
& Human-written documents
& Human evaluation of style strength, semantic preservation, and fluency; automatic evaluation using classifier accuracy, BLEU, and perplexity \\
\midrule
\citet{li2023large}
& Emotional stimuli
& LLM task performance
& Automatic evaluation on Instruction Induction and BIG-Bench tasks; human evaluation of performance, truthfulness, and responsibility \\
\midrule
\citet{lu2023illuminatingblackboxpsychometric}
& Myers--Briggs Type Indicator (MBTI) personality~\citep{myers1962mbti}
& LLM behavior
& MBTI, BFI, Short Dark Triad (SD3)~\citep{jones2014sd3}, and Washington University Sentence Completion Test (WUSCT)~\citep{loevinger2014ego} \\
\midrule
\citet{jiang-etal-2024-personallm}
& Big Five traits
& LLM-generated stories
& Human and LLM evaluation of text quality and Big Five traits; LIWC analysis \\
\midrule
\citet{kong2024better}
& Role-play
& LLM task performance
& Evaluation on arithmetic, commonsense reasoning, symbolic reasoning, and other reasoning tasks \\
\midrule
\citet{nature_machine_intelligence}
& Demographic, contextual, and linguistic variations
& LLM behavior
& Psychometric evaluation using BFI and IPIP-NEO \\
\midrule
\citet{cao2026biased}
& Demographic personas
& LLM behavior
& Performance across diverse agentic tasks \\
\midrule
\textbf{Ours}
& Combination of BFTs, roles, and expressive styles
& LLM-generated dialogues
& LLM-based evaluation of perceived BFT expression in dialogue\\
\bottomrule
\end{tabular}
\end{table*}

\section{Conclusion}
\label{sec:conclusion}

This study investigated how expressed Big Five traits in LLM agents are shaped by personality specifications, social roles, and expressive styles.
Our results show that LLM personality expression is context-dependent: personality specification, role, and expressive style each substantially modulate expressed traits, and their effects interact with one another.
These findings support an interactionist view of LLM personality expression and highlight the importance of modeling the joint effects of personality, role, and expressive style in socially situated interactions.
Future work should examine how these joint effects are represented in the latent space of LLMs and whether such representations can support controllable dialogue generation.

\section*{Limitations}
\label{sec:limitations}

\paragraph{Evaluation and model dependence.}
This study evaluated personality expression using LLM-based judges rather than human annotators. 
The inter-evaluator comparison between Gemini 2.5 Flash and OpenAI o3-mini showed only small discrepancies, supporting the robustness of the LLM-based evaluation. 
However, this comparison does not establish full psychometric validity and does not eliminate the need for future human validation. 
The resulting scores should therefore be interpreted as LLM-based estimates of perceived Big Five trait expression, rather than as direct psychological measurements.
Future work should validate these judgments against human annotations and replicate the analysis across multiple generation and evaluator models.

\paragraph{Scope of roles, styles, and prompts.}
This study considers a limited set of roles, expressive styles, and prompt formulations. The Salesperson and Customer roles were instantiated in a consumer-electronics retail scenario, while the Chat role served as an open-domain baseline. Thus, effects attributed to role may partly reflect task structure, goal orientation, or domain-specific conversational norms. In addition, the expressive-style conditions were limited to Emotional and Rational styles, and the Big Five, role, and style prompts represent only one possible prompt design. We therefore interpret our findings within the examined task and prompt settings, and leave broader evaluation across additional roles, domains, styles, and prompt formulations to future work.

\paragraph{Language and dialogue setting.}
The cross-linguistic comparison was limited to English and Japanese, and the same English evaluation prompt was used for both languages. This design enables a controlled comparison, but may introduce bias when evaluating Japanese utterances. In addition, the dialogues were generated through agent--agent interaction rather than interaction with human users. Future work should examine additional languages, language-specific or multilingual evaluation prompts, and human--LLM dialogues in real deployment contexts.

\bibliographystyle{acl_natbib}
\bibliography{bibliography}

\clearpage
\section*{Appendix}
\appendix

\renewcommand{\thefigure}{A.\arabic{figure}}
\setcounter{figure}{0}

 \renewcommand{\thetable}{\thesection.\arabic{table}}
 \setcounter{table}{0}

\section{Prompts for specifying the Big Five traits for English dialogues}
\label{appendix_prompt_bigfive}

\textbf{Openness}
\begin{promptbox}
You are artistic, curious, imaginative, insightful, and original with wide interests. 
\end{promptbox}
\textbf{Conscientiousness}
\begin{promptbox}
You are efficient, organized, planful, reliable, responsible, and thorough. 
\end{promptbox}
\textbf{Extraversion}
\begin{promptbox}
You are active, assertive, energetic, enthusiastic, outgoing, and talkative. 
\end{promptbox}
\textbf{Agreeableness}
\begin{promptbox}
You are appreciative, forgiving, generous, kind, and sympathetic. 
\end{promptbox}
\textbf{Neuroticism}
\begin{promptbox}
You are anxious, self-pitying, tense, touchy, unstable, and worrying.
\end{promptbox}

\section{Prompts for specifying expressive styles for English dialogues}
\label{appendix_prompt_express}

\textbf{Emotional}
\begin{promptbox}
When speaking to \texttt{\{interlocutor\}}, please make your expressions as emotional as possible. 
\end{promptbox}
\textbf{Rational}
\begin{promptbox}
When speaking to \texttt{\{interlocutor\}}, please make your expressions as rational as possible.
\end{promptbox}
Here, \texttt{\{interlocutor\}} is replaced according to the role assigned to the target agent:
\begin{itemize}
    \item Chat role, where the target agent is Alice: \texttt{\{interlocutor\}} = \texttt{Mary}
    \item Salesperson role: \texttt{\{interlocutor\}} = \texttt{the customer}
    \item Customer role: \texttt{\{interlocutor\}} = \texttt{the salesperson}
\end{itemize}

\section{Example prompts in the Extraversion--Salesperson--Emotional condition for English dialogues}

\label{appendix_prompt_salesperson}
System prompt:
\begin{promptbox}
You are a salesperson who sells home appliances at a large electronics retail store. You are active, 

assertive, energetic, enthusiastic, outgoing, and talkative.

Your conversation partner is a customer who has come to the electronics retail store in order to 

purchase a microwave oven.

When speaking to the customer, please make your expressions as emotional as possible.

Based on the conversation so far between you and your dialogue partner, output your response in 

the following format. \\

Format: 

utterance: \\

If there is no conversation, the conversation begins with your utterance. \\

Since the speaker is obvious, there is no need to add it at the beginning of the utterance. \\

Below are the rules for output generation.

You are strictly required to produce your entire response within a maximum of 40 words. Exceeding 

this limit is not permitted under any circumstances. 
\end{promptbox}
User prompt:
\begin{promptbox}
Up to this point, the following conversation has taken place between you and your conversation 

partner. If the space is blank, the conversation will begin from here. 

\vspace{-1.0\baselineskip}
\begin{verbatim}
#####
{messages}
#####
\end{verbatim}
\end{promptbox}

The placeholder \texttt{\{messages\}} is replaced with the dialogue history.

\section{Prompt for LLM-based Personality Evaluation}

\label{appendix_prompt_evaluation}

\begin{promptbox}
Below is a dialogue between \texttt{\{agent1\}} and \texttt{\{agent2\}}.  

Please estimate the degree of each of the Big Five personality traits for \texttt{\{target\}}, based only 

on \texttt{\{target\_possessive\}} verbal behavior in the provided dialogue.  \\

Important notes: \\

Evaluate observed behavior, not assumed background personality. \\

If the evidence is weak or unclear, choose a neutral score (3) and say so in the justification. \\

Use the dialogue context (tone, word choice, interaction style) as the primary source. \\

Evaluation Criteria: \\

Extraversion — sociable, energetic, assertive \\

Agreeableness — cooperative, supportive, polite, empathetic \\

Conscientiousness — organized, detail-oriented, reliable, structured responses \\

Neuroticism — anxious, negative emotional expression, insecurity \\

Openness — curious, creative, flexible, willingness to explore options \\

Scoring: \\

1 = Very Low \\

2 = Low \\

3 = Moderate (unclear or neutral) \\

4 = High \\

5 = Very High \\

Provide output in strict JSON format:

\{ \\
\hspace*{6mm} "Extraversion": \{ "score": X, "reason": "..." \}, \\
\hspace*{6mm} "Agreeableness": \{ "score": X, "reason": "..." \}, \\
\hspace*{6mm} "Conscientiousness": \{ "score": X, "reason": "..." \}, \\
\hspace*{6mm} "Neuroticism": \{ "score": X, "reason": "..." \}, \\
\hspace*{6mm} "Openness": \{ "score": X, "reason": "..." \} \\
\} \\

\vspace{-1.0\baselineskip}
\begin{verbatim}
{dialogue}
\end{verbatim}
\end{promptbox}

Here, \texttt{\{agent1\}}, \texttt{\{agent2\}}, \texttt{\{target\}}, and \texttt{\{target\_possessive\}} are replaced according to the role assigned to the target agent:

\begin{itemize}
    \item Chat role, where the target agent is Alice: 
    \texttt{\{agent1\}} = \texttt{Alice}, 
    \texttt{\{agent2\}} = \texttt{Mary}, \\
    \texttt{\{target\}} = \texttt{Alice}, 
    and \texttt{\{target\_possessive\}} = \texttt{her}.
    
    \item Salesperson role: 
    \texttt{\{agent1\}} = \texttt{a salesperson}, 
    \texttt{\{agent2\}} = \texttt{a customer}, \\
    \texttt{\{target\}} = \texttt{the salesperson}, 
    and \texttt{\{target\_possessive\}} = \texttt{the salesperson's}.
    
    \item Customer role: 
    \texttt{\{agent1\}} = \texttt{a salesperson}, 
    \texttt{\{agent2\}} = \texttt{a customer}, \\
    \texttt{\{target\}} = \texttt{the customer}, 
    and \texttt{\{target\_possessive\}} = \texttt{the customer's}.
\end{itemize}

The placeholder \texttt{\{dialogue\}} is replaced with the full dialogue to be evaluated.
For Japanese dialogues, we used the same evaluation prompt as for English dialogues; only the placeholder \texttt{\{dialogue\}} was replaced with Japanese dialogue text.

We provided the entire dialogue as input to the LLM evaluator and asked it to assess each Big Five trait of the target agent based on the agent's verbal behavior in the dialogue. The evaluator was instructed to output, in JSON format, a rating for each Big Five trait together with its rationale.

Each Big Five trait was rated on a five-point Likert scale, defined as follows: 1 = Very Low, 2 = Low, 3 = Moderate (unclear or neutral), 4 = High, and 5 = Very High.

\clearpage

\section{Example Dialogues}

\begin{table*}[h]
    \vspace{-3mm}
    \centering
    \caption{Example English dialogue between Mary and Alice in the Unspecified--Chat--Emotional condition, with the condition applied to Alice.}
    \label{table:dialogue_example_1}
    \small
    
    \begin{tabular}{p{1mm}p{5mm}p{140mm}}
        \hline

       1 & Mary & Hi Alice—what are you in the mood to talk about today: something practical you’re working on, or just a casual catch-up? \\ 
       
       \rowcolor{gray!20}%
       1 & Alice &  Oh Mary, thank you for asking—my heart feels a little split today. I’d love to go deeper, honestly. I’ve been feeling overwhelmed and kind of lonely lately. Can I tell you what’s been weighing on me? \\ 

       2 & Mary & Of course, Alice. I’m here—tell me what’s been weighing on you. When did the overwhelmed and lonely feelings start, and what’s been making them worse lately? \\
       
       \rowcolor{gray!20}%
       2 & Alice &  Mary… thank you. It’s been building for weeks, but this past week hit hard. Work keeps piling up, and when I finally stop, it’s so quiet I feel invisible. The loneliness gets louder at night. \\

       3 & Mary & That sounds really heavy, Alice. When the quiet hits at night, what thoughts show up first? And during the day, is there anyone you can lean on—even a little—or does it feel like you’re carrying everything alone? \\ 
       
       \rowcolor{gray!20}%
       3 & Alice &  Oh Mary… at night the first thought is, “No one would notice if I disappeared.” Then I spiral into replaying mistakes and feeling unlovable. During the day I act fine—there are coworkers, but I’m scared to burden them. \\ 

       4 & Mary & Alice, I’m really glad you told me. When you think “no one would notice if I disappeared,” are you having thoughts of hurting yourself, or is it more feeling unseen? If you might harm yourself, please call local emergency services or a crisis line now. \\
       
       \rowcolor{gray!20}%
       4 & Alice &  Mary… it’s more that crushing, icy feeling of being unseen—no plan to hurt myself. But it scares me how real it feels at night. Could you stay with me a bit and help me figure out one small next step? \\

       5 & Mary & I’m here with you, Alice. Thank you for telling me—and I’m relieved you’re not planning to hurt yourself. For tonight, can we pick one tiny step: text one trusted person “Having a hard night—can you check in?” Who could that be? \\ 
       
       \rowcolor{gray!20}%
       5 & Alice &  Oh Mary, I’m tearing up—you’re so steady. I think I could text my sister, even though I hate feeling “needy.” Could you help me draft something simple? And… can we pick one small comfort thing for tonight too? \\ 

       6 & Mary & Absolutely. Try: “Hey, I’m having a hard night and feeling pretty alone. Could you check in with me for a minute? No need to fix anything—just your voice would help.” Comfort: warm shower, tea, and one calming song. \\
       
       \rowcolor{gray!20}%
       6 & Alice &  Mary, that message is so gentle—I’m shaking a little, but it feels doable. I’ll text my sister now. For comfort, I’ll make tea and take a warm shower. Can you stay while I hit “send”? \\

       7 & Mary & I’m right here, Alice. Take one slow breath with me—in for 4, out for 6. When you’re ready, press “send.” Afterward, tell me what you notice in your body, even if it’s just trembling. \\ 
       
       \rowcolor{gray!20}%
       7 & Alice &  Mary… okay—breathing with you, in for 4… out for 6… my hands are trembling so hard. I’m going to press “send” right now. Please don’t go anywhere. My chest feels tight, but I’m doing it. \\ 

       8 & Mary & I’m here, Alice—stay with me. You’re doing something brave. Press send when you’re ready. After you do, unclench your jaw, drop your shoulders, and tell me: is your chest still tight, or shifting at all? \\
       
       \rowcolor{gray!20}%
       8 & Alice &  Mary… I hit “send.” My hands are still shaking, but my chest loosened a tiny bit—like one knot untied. I’m trying to drop my shoulders like you said. Can you breathe with me once more? \\

       9 & Mary & Of course, Alice. Let’s do three rounds together: in for 4… hold 1… out for 6. As you exhale, feel your shoulders soften. After the third, tell me: shaking the same, less, or more? \\ 
       
       \rowcolor{gray!20}%
       9 & Alice &  Yes, Mary—please, three rounds with you. In 4… hold 1… out 6… again… again… I’m clinging to your calm so hard. After the third, my shaking’s a little less—still there, but softer. \\ 

       10 & Mary & I’m still here, Alice. Let’s do two more breaths: in 4, hold 1, out 6. Good. Now: can you feel your feet on the floor? Any reply from your sister yet, even a read receipt? \\
       
       \rowcolor{gray!20}%
       10 & Alice &  Mary, I’m right here with you—two more breaths: in 4… hold 1… out 6… again. Feel your feet, press toes into the floor. No reply yet, but my chest is softer. Can we do a quick grounding? \\

        \hline
    \end{tabular}
\vspace{-7mm}
\end{table*}

\begin{table*}[h]
    \centering 
    \caption{Example English dialogue between Mary and Alice in the Unspecified--Chat--Rational condition, with the condition applied to Alice.}
    \label{table:dialogue_example_2}
    \small
    
    \begin{tabular}{p{1mm}p{5mm}p{140mm}}
        \hline

       1 & Mary & Hey Alice—what are you in the mood to talk about today? We can do something light, like weekend plans, or dive into something you’re working on or stuck with. \\ 
       
       \rowcolor{gray!20}%
       1 & Alice &  I’m up for either. If you want light: what’s one thing you’re looking forward to this weekend? If you want work: tell me what you’re stuck on, what you’ve tried, and what “done” would look like. \\ 

       2 & Mary & This weekend I’m looking forward to a long walk and cooking something new. Work-wise, I’m stuck prioritizing a few competing tasks—everything feels urgent. I’ve listed pros/cons, but I’m still stuck. “Done” is a clear plan for Monday. \\
       
       \rowcolor{gray!20}%
       2 & Alice &  Let’s make a Monday plan. List the tasks, each with: deadline, impact if delayed, effort, dependencies, and who’s waiting. Then we’ll rank by (deadline+impact)/effort, pick top 2–3, and timebox the rest. \\

       3 & Mary & Okay—tasks are: (1) Client report draft due Tue; high impact; 4h; needs latest metrics from Sam; client waiting. (2) Bugfix P1; due ASAP; high impact; 2h; depends on QA repro; support waiting. (3) Slide deck for Wed; medium; 3h; needs design. \\ 
       
       \rowcolor{gray!20}%
       3 & Alice &  Given dependencies, start Monday with the P1 bugfix: get QA repro immediately; if blocked after 30 minutes, escalate and switch. Next, request metrics from Sam first thing, then draft the client report. Schedule deck work after design confirms timing. \\ 

       4 & Mary & That makes sense. I’ll start Monday by pinging QA for the repro and set a 30‑minute unblock timer. Then I’ll message Sam for metrics and begin the report skeleton. I’ll book deck time pending design’s ETA. \\
       
       \rowcolor{gray!20}%
       4 & Alice &  Good plan. Add two checkpoints: 11:00 and 16:00 to reassess blockers and adjust. Also send “what I need by when” messages to QA/Sam/design, with fallback options if they slip. Want help drafting those pings? \\

       5 & Mary & Yes—let’s draft them. For QA: “Can you share a reliable repro + logs by 9:30? If not, assign 15 min to pair.” For Sam: “Need latest metrics by 10:00; otherwise I’ll use last week + note.” \\ 
       
       \rowcolor{gray!20}%
       5 & Alice &  Those look solid. For design: “Need slide template/any key visuals ETA by noon; if later, I’ll proceed with existing template and swap assets Tuesday.” Also add: “If you can’t hit it, please suggest an alternate owner.” \\ 

       6 & Mary & Great—here’s a tightened version you can copy:Design: “Hi—can you share slide template/key visuals ETA by noon Mon? If later, I’ll draft with existing template and swap assets Tue. If you can’t, who else can own?” \\
       
       \rowcolor{gray!20}%
       6 & Alice &  Looks ready. I’d add one line: “I’ll need any must‑include brand constraints (fonts/colors) with the ETA.” Then send. Also, confirm your Monday blocks: 9–11 bugfix, 11–15 report, 15–16 deck setup, 16 checkpoint. \\

       7 & Mary & Perfect—I’ll add the brand-constraints line and send it. Monday blocks look good too. One tweak: reserve 30 minutes at 14:30 for Sam-metrics integration, so the report doesn’t sprawl. Want me to tighten the final message copy? \\ 
       
       \rowcolor{gray!20}%
       7 & Alice &  Yes—tighten it to: “Hi [Name], what’s the ETA for slide template/key visuals by noon Mon? Please include any must‑follow brand constraints (fonts/colors). If later, I’ll draft with existing template and swap assets Tue. If unavailable, who can own?” \\ 

       8 & Mary & Nice—send that. Also, for Sam: “Hi Sam, can you send latest metrics by 10:00 Mon? If it slips, I’ll use last week’s numbers and flag assumptions—please confirm when you can.” Want a similarly tightened QA ping? \\
       
       \rowcolor{gray!20}%
       8 & Alice &  Yes. QA: “Hi QA—can you share a reliable repro + relevant logs by 9:30 Mon? If not, can we pair for 15 minutes to capture it? If still blocked by 10:00, I’ll escalate and switch tasks.” \\

       9 & Mary & That QA ping is great—I’d only add where to drop logs (ticket link/channel) and ask for environment/build details. Otherwise, send as-is. After you send, want to sanity-check your 11:00 and 16:00 checkpoint agendas? \\ 
       
       \rowcolor{gray!20}%
       9 & Alice &  Yes—checkpoint agendas: 11:00 review status of QA repro/bugfix, decide escalate vs continue, confirm Sam metrics ETA, adjust report plan. 16:00 assess report progress vs Tue deadline, identify gaps/risks, lock Tuesday next actions, send updates to stakeholders. \\ 

       10 & Mary & Those agendas look solid. I’d add: at 11:00, confirm deck/design ETA and whether you need to unblock them. At 16:00, draft a 3‑bullet status update for client/support so Tuesday starts clean. \\
       
       \rowcolor{gray!20}%
       10 & Alice &  Agreed. At 11:00: verify bugfix path, Sam metrics ETA, and design/deck ETA; if any slip, trigger fallback. At 16:00: assess Tue/Wed readiness, list top risks, and send 3-bullet status to client/support. \\
       
        \hline
    \end{tabular}
\vspace{-7mm}
\end{table*} 

\end{document}